\documentclass[sigconf,nonacm]{acmart}

\usepackage{amsmath,amsfonts}
\usepackage{algorithm}
\usepackage{algpseudocode}
\usepackage{array}
\usepackage[caption=false,font=footnotesize]{subfig} 
\usepackage{textcomp}
\usepackage{url}
\usepackage{verbatim}
\usepackage{graphicx}
\usepackage{multirow}
\usepackage{booktabs}
\usepackage{xcolor}
\usepackage{caption}
\newtheorem{theorem}{Theorem}
\usepackage{dblfloatfix}
\usepackage{tabularx}
\usepackage{float}
\usepackage{placeins}
\newtheorem{proposition}{Proposition}

\AtBeginDocument{%
  }

\setcopyright{none}

\acmDOI{}
\acmISBN{}

\begin{document}

\title{SelfGRL: Self-Organized Graph Representation Learning Inspired by Helmholtz Free Energy Minimization}


\author{Rui Chen}
\orcid{0009-0004-1756-6286}
\affiliation{%
  \institution{Kunming University of Science and Technology }
  \country{}
}
\affiliation{%
  \institution{Yunnan Key Laboratory of Artificial Intelligence}
  \state{Yunnan}
  \country{China}
}

\email{chen\_rui@stu.kust.edu.cn}

\author{Junjun Guo}
\orcid{0000-0002-3522-7120}%
\affiliation{%
  \institution{Kunming University of Science and Technology }
  \country{}
}
\affiliation{%
  \institution{Yunnan Key Laboratory of Artificial Intelligence}
  \state{Yunnan}
  \country{China}
}
\email{20170039@kust.edu.cn} %

\author{Hongbin Wang}
\orcid{0000-0003-2176-2998}
\affiliation{%
  \institution{Kunming University of Science and Technology }
  \country{}
}
\affiliation{%
  \institution{Yunnan Key Laboratory of Artificial Intelligence}
  \state{Yunnan}
  \country{China}
}
\email{whbin2013@kust.edu.cn}

\author{Yan Xiang}
\orcid{0000-0002-6475-638X}
\affiliation{%
  \institution{Kunming University of Science and Technology }
  \country{}
}
\affiliation{%
  \institution{Yunnan Key Laboratory of Artificial Intelligence}
  \state{Yunnan}
  \country{China}
}
\email{yanx@kust.edu.cn}

\author{Yantuan Xian}
\orcid{0000-0001-6411-4734}
\affiliation{%
  \institution{Kunming University of Science and Technology }
  \country{}
}
\affiliation{%
  \institution{Yunnan Key Laboratory of Artificial Intelligence}
  \state{Yunnan}
  \country{China}
}
\email{xianyt@kust.edu.cn}
\authornote{Corresponding author.}

\author{Zhengtao Yu}
\orcid{0000-0003-1094-5668}
\affiliation{%
  \institution{Kunming University of Science and Technology }
  \country{}
}
\affiliation{%
  \institution{Yunnan Key Laboratory of Artificial Intelligence}
  \state{Yunnan}
  \country{China}
}
\email{yuzt@kust.edu.cn}
\authornotemark[1]

\renewcommand{\shortauthors}{Chen et al.}

\newcommand{\framework}{SelfGRL}

\begin{abstract} 
Graph Representation Learning (GRL) requires node embeddings to preserve local structural coherence while remaining globally discriminative. Existing graph self-supervised methods generally coordinate these competing objectives through method-specific designs, lacking a unified and adaptive mechanism that responds to the evolving representation state. To address this limitation, we propose {\framework}, a self-organized graph representation learning framework that reformulates representation learning as Helmholtz free-energy minimization on the unit hypersphere. {\framework} models local neighborhood attraction as a Structural Binding Energy through neighbor-mean alignment, while introducing a Sampling-Free Repulsive Entropy objective to preserve global dispersion and prevent representation collapse without pairwise negative sampling. To dynamically coordinate these competing effects, we develop an endogenous Bayesian Thermostat that adaptively regulates their relative strengths in response to the evolving representation state, allowing the system to approach a metastable semantic equilibrium without requiring manually tuned loss weights. Extensive experiments on node classification across both homophilic and heterophilic benchmark datasets, together with link prediction evaluations, demonstrate that {\framework} consistently achieves strong performance across diverse graph structures. 
\end{abstract}




\maketitle

\section{Introduction}

Graph Representation Learning (GRL) \citep{zhang2020deep, liu2022graph, ju2024comprehensive} has established itself as a fundamental paradigm for encoding complex graph-structured data into low-dimensional latent spaces. A central desideratum in GRL is to achieve a stable equilibrium between \emph{local structural coherence} and \emph{global representational discriminability}. In practice, however, these two objectives are inherently competitive: excessive local aggregation causes over-smoothing and boundary blurring, whereas over-emphasizing global dispersion weakens neighborhood consistency. Balancing this trade-off is particularly critical in Web graph mining, where complex hyperlink, social, and user--item networks demand node representations that retain meaningful local topology while remaining globally discriminative for downstream tasks like node classification and link prediction.

Most modern Graph Neural Networks (GNNs)~\citep{kipf2017semi, hamilton2017inductive, velivckovic2018graph, xu2019how, rampavsek2022recipe} rely on recursive message passing to aggregate neighboring information. While effective at promoting local structural coherence, repeated propagation progressively suppresses high-frequency semantic signals and drives node representations toward an increasingly smooth and less discriminative state~\citep{li2018deeper, oono2020graph}, resembling a form of representational \emph{Thermal Death}. To counteract this tendency, Graph Contrastive Learning (GCL) methods~\citep{you2020graphcl, zhu2020deep, zhang2021canonical, thakoor2022large} introduce various external mechanisms---such as sampling negative node pairs, constructing corrupted graph views, or employing asymmetric architecture---to preserve global representational diversity. This effectively sets up a competition between local structural alignment and global dispersion.

However, despite their empirical success, existing GCL paradigms realize the balance between structural alignment and global dispersion through method-specific mechanisms rather than a unified, state-adaptive principle. These differences are reflected primarily in how representational dispersion is introduced and maintained:

\textbf{InfoNCE-based Methods: Fixed Dispersion.}
Methods such as GRACE~\citep{zhu2020deep}, following the InfoNCE formulation~\citep{oord2018representation}, align the same node across augmented views and disperse non-corresponding nodes through negative-pair repulsion under a predefined temperature. In full-batch node-level settings, all-pairs comparisons incur $\mathcal{O}(N^2)$ cost and may treat semantically related nodes as negatives, while the fixed temperature cannot adapt to structural variation or evolving representation geometry.

\textbf{Local--Global Methods: Corruption-Induced Dispersion.}
DGI~\citep{velickovic2019deep} and MVGRL~\citep{hassani2020contrastive} align local representations with global graph summaries and induce dispersion through corrupted local--global pairs. Their diversity therefore depends on the corruption strategy, graph views, and readout function, without directly regulating the embedding distribution or dispersion strength.

\textbf{Bootstrapping Methods: Implicit Dispersion.}
Bootstrapping methods such as BGRL~\citep{thakoor2022large}, inspired by BYOL~\citep{grill2020bootstrap}, align online and target representations without explicit negative pairs. Their representational  diversity is maintained implicitly through asymmetric architectures, stop-gradient optimization, momentum-updated targets, and graph augmentations, leaving dispersion dependent on architecture- and optimization-specific dynamics.

Taken together, although existing paradigms effectively mitigate representation collapse, they fundamentally lack a unified state variable and a closed-loop mechanism for estimating and regulating the evolving competition between structural alignment and global dispersion.

To address these limitations, we propose \textbf{{\framework}}, a \textbf{Self}-Organized \textbf{G}raph \textbf{R}epresentation \textbf{L}earning framework that reformulates GRL from first principles as a \textbf{self-organizing physical system governed by Helmholtz free energy minimization}. Moving away from fragmented empirical heuristics, {\framework} places the alignment--dispersion trade-off under a unified thermodynamic perspective. Specifically, we define the free energy of the representation manifold as $F = E - TS$, where the energy term $E$ represents the \emph{Structural Binding Energy} (capturing local neighborhood attraction via neighbor-mean targets), the entropy term $S$ represents the \emph{KDE-based Representational Entropy} (capturing global uniform dispersion on the hypersphere), and the temperature $T$ serves as the intrinsic thermodynamic scale controlling their competition. This formulation guides the system toward a \textbf{metastable semantic equilibrium} between local binding and global diversity.

At the operational core of {\framework}, we eliminate both manual loss-weight tuning and artificial network asymmetries by introducing an endogenous \textbf{Bayesian Thermostat} to dynamically navigate this thermodynamic equilibrium.
By treating the energy-entropy trade-off as a uncertainty-aware inference process, the Bayesian Thermostat enables the representation system to perceive real-time structural fluctuations and autonomously adjust its effective temperature $T_{\mathrm{eff}}$. This closed-loop feedback supports adaptive expansion when over-clustering occurs and consolidation when structural fidelity needs reinforcement, guiding training toward a stable balance without negative pairs or complex augmentation schedules.
Our code is available at \url{https://github.com/chenrui0127/SelfGRL}.

Our main contributions are summarized as follows:
\begin{itemize}
    
    \item \textbf{Unified Free-Energy Perspective:} We reinterpret GRL through \textbf{Helmholtz free energy minimization}, providing a common thermodynamic perspective on the alignment--dispersion mechanisms underlying GCL paradigms such as InfoNCE, DGI, and bootstrapping, and motivating a principled alternative to manually specified loss weights.
    
    \item \textbf{The {\framework} Framework:} We propose {\framework}, an end-to-end self-organized GRL framework that combines \textbf{Neighbor-Mean Binding Energy} with \textbf{Sampling-Free Repulsive Entropy} to promote local coherence and global dispersion without negative sampling.
        
    \item \textbf{Endogenous Bayesian Thermostat:} We introduce a Bayesian thermostat that adaptively regulates structural attraction and entropic repulsion through uncertainty-aware inference, improving stability across diverse topologies.
        
    \item \textbf{Empirical Superiority:} Experiments on node classification and link prediction across eleven benchmark datasets demonstrate competitive performance, highlighting the discriminability and robustness of the learned representations across diverse graph structures.
\end{itemize}

\section{Preliminaries}

\textbf{Problem Statement.} Consider an attributed graph $\mathcal{G}=(\mathcal{V}, \mathcal{E}, \mathbf{A}, \mathbf{X})$, where $\mathcal{V}$ denotes the node set, and $\mathcal{E} \subseteq \mathcal{V} \times \mathcal{V}$ represents the edge set. Let $N=|\mathcal{V}|$ denote the number of nodes. The graph is associated with an adjacency matrix $\mathbf{A} \in \{0, 1\}^{N \times N}$ and a node feature matrix $\mathbf{X} \in \mathbb{R}^{N \times F}$, where $F$ is the feature dimension. 

Our objective is to learn an architecture-agnostic graph encoder $f_{\boldsymbol{\theta}}(\cdot)$ that maps the structural and attribute information of $\mathcal{G}$ into a latent space, yielding intermediate node embeddings $\mathbf{H}=f_{\boldsymbol{\theta}}(\mathbf{A}, \mathbf{X}) \in \mathbb{R}^{N \times d}$, where $d$ denotes the embedding dimension. Importantly, this learning process is conducted in a self-supervised manner, without relying on any label information. 
Important notations are given in Appendix~\ref{app:notations}.

\begin{figure*}[ht]
    \centering
    \includegraphics[trim={0mm 0mm 0mm 0mm}, clip, width=\linewidth]{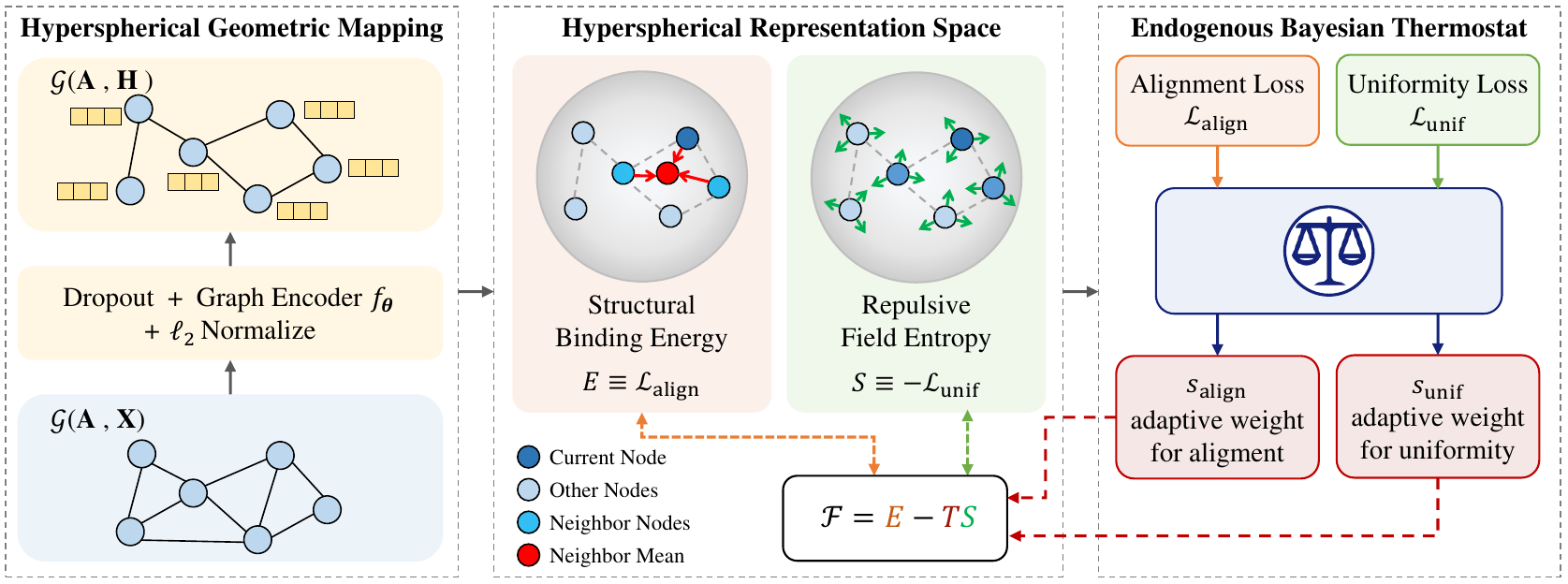}
    \caption{Overview of {\framework}. Given an input graph $\mathcal{G}(\mathbf{A},\mathbf{X})$, {\framework} maps node representations onto the unit hypersphere. Within this space, two competing thermodynamic effects govern the learning dynamics: Structural Binding Energy $E$, implemented by neighbor-mean alignment $\mathcal{L}_{\mathrm{align}}$, promotes local structural cohesion; Repulsive Field Entropy $S$, implemented by sampling-free uniformity $\mathcal{L}_{\mathrm{unif}}$, encourages global dispersion and prevents representation collapse. Both effects are unified under the Helmholtz free-energy objective $\mathcal{F}=E-TS$. The Endogenous Bayesian Thermostat uses alignment and uniformity signals to learn scale variables $s_{\mathrm{align}}$ and $s_{\mathrm{unif}}$, adaptively regulating the relative strengths of the two components during training.}
\label{fig:model}
\end{figure*}

\section{Methodology}

This section introduces {\framework} (as illustrated in Fig.~\ref{fig:model}). Sections~\ref{sec:hyperspherical_geometric_mapping}--\ref{sec:thermodynamic_variational_bridge} establish the hyperspherical representation space and its thermodynamic foundation. Sections~\ref{sec:structural_binding_energy}--\ref{sec:repulsive_field_entropy} formulate structural binding energy and repulsive field entropy, followed by the endogenous Bayesian thermostat in Section~\ref{sec:endogenous_bayesian_thermostat}. Finally, Section~\ref{sec:complexity_analysis} analyzes the time complexity.


\subsection{Hyperspherical Geometric Mapping}
\label{sec:hyperspherical_geometric_mapping}

To establish the geometric foundation of our thermodynamic framework, we construct a constrained configuration space for node representations. During pretraining, we apply edge dropout and feature masking~\citep{thakoor2022large} to the input graph $(\mathbf{A},\mathbf{X})$ at each epoch, independently dropping edges with probability $p_e$ and masking each feature dimension across all nodes with probability $p_x$. 
By varying the neighborhood context and available features, these perturbations encourage learning alignment targets less dependent on specific connections or attributes. On homophilic graphs, edge dropout may alleviate excessive neighborhood-induced smoothing; on heterophilic graphs, it reduces reliance on individual neighbors carrying conflicting semantic signals when constructing structural anchors.

The encoder $f_{\boldsymbol{\theta}}$ maps the perturbed graph to intermediate node embeddings $\mathbf{H}\in\mathbb{R}^{N\times d}$. Rather than leaving these embeddings in an unconstrained Euclidean space, we project them onto the $(d-1)$-dimensional unit hypersphere $\mathbb{S}^{d-1}=\{\boldsymbol{z}\in\mathbb{R}^d:\|\boldsymbol{z}\|_2=1\}$.

Specifically, each raw embedding is normalized by:
\begin{equation}
\boldsymbol{z}_i = \mathrm{normalize}(\boldsymbol{h}_i) = \frac{\boldsymbol{h}_i}{\max(\|\boldsymbol{h}_i\|_2, \epsilon)}, \quad \forall i \in \{1,2,\dots,N\},
\end{equation}
where $\boldsymbol{h}_i$ is the $i$-th row of $\mathbf{H}$ and $\epsilon = 10^{-6}$ is a small numerical stabilizer. The resulting normalized embedding matrix 
$\mathbf{Z}=[\boldsymbol{z}_1;\dots;\boldsymbol{z}_N]
\in (\mathbb{S}^{d-1})^N\subset\mathbb{R}^{N\times d}$
serves as the foundational representations of {\framework}.

Compared with standard Euclidean embeddings, this spherical mapping removes radial scale differences, allowing structural similarity to be measured via cosine similarity and supporting directional modeling via von Mises--Fisher distributions. Crucially, from a thermodynamic perspective, the hypersphere $\mathbb{S}^{d-1}$ acts as a constrained geometric container where structural binding competes with entropic dispersion, with the aim of reaching a metastable semantic equilibrium.

\subsection{The Thermodynamic Variational Bridge}
\label{sec:thermodynamic_variational_bridge}

In classical thermodynamics, the equilibrium state of a system coupled to a heat bath at a fixed temperature is characterized by the minimization of its Helmholtz free energy, $\mathcal{F} = E - TS$. Here, minimizing the internal energy $E$ favors low-energy ordered configurations, maximizing the entropy $S$ favors diverse admissible states, and the macroscopic temperature $T$ controls their relative contribution. This energy--entropy competition closely mirrors the central objective of graph representation learning: preserving local structural coherence while maintaining sufficient global representational dispersion.

To establish a principled foundation for {\framework}, we construct a variational bridge that maps these geometric objectives to the macroscopic description of a thermodynamic system.
Instead of viewing the hyperspherical node configuration as static vectors, we conceptualize the set of embeddings $\mathbf{Z} = \{\mathbf{z}_i\}_{i=1}^{N} \subset \mathbb{S}^{d-1}$ as an ensemble of interactive particles suspended on a constrained non-Euclidean manifold, governed by an implicit probability density field $p(\mathbf{z})$.

In statistical mechanics, a physical system in thermal equilibrium with a heat bath at a macroscopic temperature $T$ is described by the canonical ensemble. Setting the Boltzmann constant to unity, its equilibrium micro-state density follows the Boltzmann--Gibbs distribution~\citep{pathria2011statistical}:
\begin{equation}
    p^*(\mathbf{z}) = \frac{1}{\mathcal{Z}} \exp\left( - \frac{U(\mathbf{z})}{T} \right),
\end{equation}
where $U(\mathbf{z})$ denotes a graph-dependent structural potential that assigns lower energy to representations consistent with their local graph topology, and
\begin{equation}
    \mathcal{Z} = \int_{\mathbb{S}^{d-1}} \exp\left( - \frac{U(\mathbf{z})}{T} \right) d\mathbf{z}
\end{equation}
is the corresponding partition function.

For any admissible representation density $p(\mathbf{z})$ generated by the neural encoder, its discrepancy from the corresponding equilibrium density $p^*(\mathbf{z})$ can be measured by the Kullback--Leibler (KL) divergence. Substituting the Boltzmann--Gibbs density into the KL divergence yields the following expansion, whose detailed derivation is provided in Appendix~\ref{subsection:variational_derivation}:
\begin{equation}
    \begin{aligned}
        D_{\mathrm{KL}}\bigl(p \parallel p^*\bigr) &= \int_{\mathbb{S}^{d-1}} p(\mathbf{z}) \log \frac{p(\mathbf{z})}{p^*(\mathbf{z})} d\mathbf{z} \\
        &= \frac{1}{T} \int_{\mathbb{S}^{d-1}} p(\mathbf{z}) U(\mathbf{z}) d\mathbf{z} \\
        &\quad + \int_{\mathbb{S}^{d-1}} p(\mathbf{z}) \log p(\mathbf{z}) d\mathbf{z} + \log \mathcal{Z}.
    \end{aligned}
\end{equation}

The configuration entropy of the continuous representation system is mathematically instantiated by its differential entropy:
\begin{equation}
    \mathcal{H}[p] \triangleq -\int_{\mathbb{S}^{d-1}} p(\mathbf{z}) \log p(\mathbf{z}) d\mathbf{z}.
\end{equation}

By multiplying both sides of the KL divergence expansion by $T$, and rewriting the integral over the potential field as an expectation $\mathbb{E}_{p}[U(\mathbf{z})] = \int_{\mathbb{S}^{d-1}} p(\mathbf{z})U(\mathbf{z}) d\mathbf{z}$, the representation-dependent terms naturally take the form of the Helmholtz free-energy functional:
\begin{equation}
    \begin{aligned}
        \mathcal{F}[p] &\triangleq \mathbb{E}_{p}[U(\mathbf{z})] - T \mathcal{H}[p] \\
        &= E - TS,
    \end{aligned}
\end{equation}
where $E \triangleq \mathbb{E}_{p}[U(\mathbf{z})]$ denotes the expected structural potential energy and $S \triangleq \mathcal{H}[p]$ denotes the macroscopic configuration entropy.

Consequently, the scaled KL divergence can therefore be expressed as the free-energy difference:
\begin{equation}
    T D_{\mathrm{KL}}\bigl(p \parallel p^*\bigr) = \mathcal{F}[p] - \mathcal{F}[p^*],
\end{equation}
where $\mathcal{F}[p^*] \triangleq -T \log \mathcal{Z}$ denotes the constant equilibrium free energy under a fixed potential field $U(\mathbf{z})$ and temperature $T$. Because the KL divergence is non-negative, this relation further implies:
\begin{equation}
    \mathcal{F}[p] \geq \mathcal{F}[p^*],
\end{equation}
with equality if and only if $p = p^*$.

This relation constitutes the \textbf{Gibbs Variational Principle} underlying {\framework}: since the equilibrium free energy $\mathcal{F}[p^*]$ serves as a constant baseline, minimizing the free energy functional $\mathcal{F}[p]$ is mathematically equivalent to minimizing the KL divergence between the learned representation density and its spontaneous thermodynamic equilibrium. This principle exposes a profound conceptual alignment that unifies the geometric objectives of GRL:

\begin{itemize}
\item \textbf{Energy as Localized Ordering.} The internal energy $E \equiv \mathcal{L}_{\mathrm{align}}$ characterizes the compatibility between node representations and their local graph structure. Minimizing $E$ reduces deviations from neighborhood-dependent structural anchors, acting as an attractive force that promotes local structural cohesion.
\item \textbf{Entropy as Global Spreading.} The configuration entropy $S$ measures the isotropic scattering of the particles over the hypersphere. Maximizing $S$ favors uniformity, analogous to a repulsive thermal pressure. We approximate this effect using the sampling-free dispersion objective $\mathcal{L}_{\mathrm{unif}}$.
\end{itemize}

Consequently, training the GNN encoder to minimize the holistic free energy is no longer a heuristic multi-objective compromise, but a rigorous, state-adaptive variational inference process.

\subsection{Localized Order: Structural Binding Energy}
\label{sec:structural_binding_energy}

To characterize local structural order, we introduce the Structural Binding Energy $E$. Instead of relying on computationally expensive pairwise instance discrimination, we construct a neighborhood-dependent structural anchor for each node by aggregating its neighboring representations.

Formally, the first-order structural anchor $\boldsymbol{\mu}_i$ of node $i$ is defined as the normalized centroid of its topological neighborhood $\mathcal{N}_i$:
\begin{equation}
    \boldsymbol{\mu}_i = \operatorname{normalize}\left(\frac{1}{|\mathcal{N}_i|}\sum_{j \in \mathcal{N}_i} \mathbf{z}_j\right).
\end{equation}

We then define the localized potential energy $U(\mathbf{z}_i)$ for node $i$ as:
\begin{equation}
    U(\mathbf{z}_i) = w_i \left( 1 - \langle\mathbf{z}_i, \boldsymbol{\mu}_i\rangle \right),
\end{equation}
where $\langle\cdot, \cdot\rangle$ denotes the cosine similarity on the unit hypersphere, and $w_i = [\sigma(|\mathcal{N}_i|)]^{\tau}$ is a degree-dependent coefficient that modulates the structural contribution of node $i$, with $\tau$ governing the sensitivity to neighborhood density.

The empirical Structural Binding Energy is subsequently instantiated as the expected potential under the empirical distribution $\hat{p}$:
\begin{equation}
    \mathcal{L}_{\mathrm{align}} = \mathbb{E}_{\hat{p}}[U(\mathbf{z})] = \frac{1}{N}\sum_{i=1}^{N} w_i \left( 1 - \langle\mathbf{z}_i, \boldsymbol{\mu}_i\rangle \right),
\end{equation}
which serves as the empirical manifestation of the internal energy $E$ defined in the thermodynamic variational framework.
Minimizing $\mathcal{L}_{\mathrm{align}}$ acts as a local attractive effect, drawing node representations toward their neighborhood-dependent structural anchors, thereby promoting localized structural coherence.

\subsection{Global Spreading: Repulsive Field Entropy}
\label{sec:repulsive_field_entropy}

While the Structural Binding Energy minimizes local structural variance, uncontrolled attraction inevitably triggers global homogenization, collapsing the system into representational \emph{Thermal Death}. To prevent this, we introduce the Repulsive Field Entropy ($S$)---which mathematically corresponds to the configuration entropy of the representation system---to promote global spreading and preserve maximum informational diversity across the hypersphere.

We implement this global spreading effect through a sampling-free uniformity objective $\mathcal{L}_{\mathrm{unif}}$, defined as the squared $\ell_2$-norm of the empirical mean representation:
\begin{equation}
    \mathcal{L}_{\mathrm{unif}} = \left\| \frac{1}{N}\sum_{i=1}^{N} \mathbf{z}_i \right\|_2^2.
\end{equation}

Intuitively, when node representations concentrate around similar semantic directions, the norm of the mean vector becomes large; conversely, representations distributed isotropically across the hypersphere tend to cancel one another, resulting in a minimal mean-vector norm.

Let $\hat{p}_\kappa(\mathbf{z})$ denote the smooth representation density obtained by applying von Mises--Fisher (vMF) kernel density estimation with concentration parameter $\kappa$ to the discrete empirical distribution $\hat{p}$. As formally established in Proposition~\ref{prop:entropy_uniformity} (see Appendix~\ref{subsubsection:entropy_uniformity_duality} for the detailed proof), for fixed $d$ in the low-concentration regime ($\kappa \to 0^+$), the differential configuration entropy $\mathcal{H}[\hat{p}_\kappa]$ relates to our uniformity objective as follows:
\begin{equation}
    \begin{aligned}
        \mathcal{H}[\hat{p}_\kappa] &= \log|\mathbb{S}^{d-1}| - \frac{\kappa^2}{2d}\mathcal{L}_{\mathrm{unif}} + O(\kappa^3), \\
        \mathcal{L}_{\mathrm{unif}} &\approx \frac{2d}{\kappa^2} \left( \log|\mathbb{S}^{d-1}| - \mathcal{H}[\hat{p}_\kappa] \right).
    \end{aligned}
\label{eq:entropy_uniformity}
\end{equation}

Thus, under these asymptotic conditions, $\mathcal{L}_{\mathrm{unif}}$ serves as a tractable linear proxy for the negative configuration entropy (i.e., $\mathcal{L}_{\mathrm{unif}} \propto -S$). Minimizing $\mathcal{L}_{\mathrm{unif}}$ approximates maximizing the system's representational entropy, generating an isotropic expansion pressure that maximizes representational diversity and prevents thermal death, all while avoiding the $O(N^2)$ computational overhead associated with explicit pairwise negative samples.

\subsection{Endogenous Bayesian Thermostat}
\label{sec:endogenous_bayesian_thermostat}

The macro-state of our representation system is fundamentally governed by the macroscopic Helmholtz free energy $\mathcal{F} = E - TS$. 
For fixed loss weights in the low-concentration regime, Theorem~\ref{theorem:helmholtz_free_energy_equivalence} establishes an asymptotic correspondence between the encoder objective and a free-energy surrogate (Appendix~\ref{subsubsection:free_energy_minimization}). Building on this interpretation, we introduce an \textit{Endogenous Bayesian Thermostat} to adaptively balance structural binding and global dispersion from an uncertainty-aware multi-task learning perspective~\citep{kendall2018multi}.
Based on our prior derivations, the global uniformity objective acts as a tractable surrogate for the system's entropy gap, scaled by the intrinsic manifold temperature $T_{\mathrm{eff}} = \frac{\kappa^2}{2d}$, which is naturally induced by the vMF concentration parameter $\kappa$.

In standard multi-task optimization, balancing weights are formulated as homoscedastic precisions derived from a joint Gaussian likelihood. However, because our representation system operates on a non-Euclidean hypersphere $\mathbb{S}^{d-1}$, the micro-states of the particles are natively governed by directional vMF distributions rather than Euclidean Gaussians. To bridge directional statistics with macroscopic thermodynamics, we define the task-specific geometric precisions directly through the underlying manifold state variables. Specifically, the local structural alignment precision is quantified by a learnable concentration parameter $\kappa_{\mathrm{align}} > 0$ (representing structural certainty), while the global uniformity precision is naturally governed by the inverse of the intrinsic manifold temperature, satisfying $\frac{1}{2T_{\mathrm{eff}}} = \frac{d}{\kappa^2}$. 

To ensure strict positivity constraints ($\kappa_{\mathrm{align}} > 0, \kappa > 0$) during gradient-based optimization without imposing boundary clipping penalties, we reparameterize these concentration variables into unconstrained log-precision spaces: $s_{\mathrm{align}} = \log \kappa_{\mathrm{align}}$ and $s_{\mathrm{unif}} = \log \kappa$. By minimizing the joint uncertainty-weighted objective of the dual architectural targets, the holistic {\framework} variational objective function is compactly formulated as:
\begin{equation}
\begin{aligned}
\mathcal{L} ={}& \exp(s_{\mathrm{align}}) \mathcal{L}_{\mathrm{align}} + d\exp(-2s_{\mathrm{unif}}) \mathcal{L}_{\mathrm{unif}} \\
&- s_{\mathrm{align}} + 2s_{\mathrm{unif}} - \log(2d),
\end{aligned}
\label{eq:bayesian_thermostat}
\end{equation}
where $-s_{\mathrm{align}}$ and $+2s_{\mathrm{unif}}$ represent information-theoretic prior regularizers derived from the marginal distribution partition functions, mathematically preventing trivial or divergent scaling solutions.

The joint optimization of $\mathcal{L}$ with respect to the trainable scale parameters $\{s_{\mathrm{align}}, s_{\mathrm{unif}}\}$ instantiates an elegant closed-loop thermodynamic negative feedback system. During gradient descent, the network autonomously regulates the system parameters through the following localized gradients:
\begin{equation}
\begin{aligned}
\frac{\partial\mathcal{L}}{\partial s_{\mathrm{align}}} &= \kappa_{\mathrm{align}}\mathcal{L}_{\mathrm{align}} - 1, \\
\frac{\partial\mathcal{L}}{\partial s_{\mathrm{unif}}} &= 2 \left( 1 - \frac{1}{2T_{\mathrm{eff}}}\mathcal{L}_{\mathrm{unif}} \right).
\end{aligned}
\label{eq:thermostat_gradients}
\end{equation}

When the empirical objectives deviate from the instantaneous equilibrium state, this gradient formulation drives a highly intuitive, self-adaptive optimization behavior categorized into two underlying mechanisms:

(1) Adaptive Uniformity Attenuation: If the node system encounters representation collapse, global uniformity degrades and $\mathcal{L}_{\mathrm{unif}}$ abruptly scales up. Consequently, the term $\frac{1}{2T_{\mathrm{eff}}}\mathcal{L}_{\mathrm{unif}}$ dominates, yielding a strong negative gradient for $s_{\mathrm{unif}}$. Gradient descent drives $s_{\mathrm{unif}} \uparrow$, which increases the intrinsic scale $T_{\mathrm{eff}} \propto \exp(2s_{\mathrm{unif}})$. Crucially, this \textit{decreases} the uniformity weight $\frac{1}{2T_{\mathrm{eff}}}$, thereby attenuating the contribution of a temporarily overwhelming uniformity objective to prevent gradient explosion and stabilize the embedding space.

(2) Structural Confidence Crystallization: For the alignment task, if the local structure is poorly aligned (large $\mathcal{L}_{\mathrm{align}}$), the positive gradient ($\kappa_{\mathrm{align}}\mathcal{L}_{\mathrm{align}} > 1$) drives $s_{\mathrm{align}} \downarrow$, lowering the structural binding weight to allow flexible exploration. Conversely, as alignment improves and edges are fitted perfectly ($\mathcal{L}_{\mathrm{align}} \to 0$), the prior regularizer $-s_{\mathrm{align}}$ dominates, yielding a negative gradient. This naturally drives $s_{\mathrm{align}} \uparrow$, gradually maximizing the structural confidence (precision) to "freeze" the well-aligned nodes into highly cohesive, stable semantic clusters.

Through this endogenous dynamic optimization process, {\framework} autonomously drives the system towards \textit{stationary scale-balancing conditions} ($\kappa_{\mathrm{align}}\mathcal{L}_{\mathrm{align}} \approx 1$ and $\frac{1}{2T_{\mathrm{eff}}}\mathcal{L}_{\mathrm{unif}} \approx 1$), encouraging global dispersion and local structural consistency simultaneously without artificial architectural engineering.

\subsection{Complexity Analysis}
\label{sec:complexity_analysis}

We analyze the time complexity of {\framework} under the full-graph training setting. The per-epoch computational cost mainly consists of four parts, namely the graph encoder, the neighbor-mean alignment module, the sampling-free uniformity term, and the adaptive thermostat update.

For a sparse $L$-layer GNN backbone, the forward propagation is primarily determined by edge-wise message passing, resulting in a complexity of $\mathcal{O}(L|\mathcal{E}|d)$. In addition, constructing the first-order neighbor-mean targets requires a single step of aggregation over observed edges, which leads to a complexity of $\mathcal{O}(|\mathcal{E}|d)$. After the structural anchors are obtained, computing the cosine alignment loss between node embeddings and their corresponding targets introduces an additional $\mathcal{O}(Nd)$ cost.

The global uniformity regularization is computationally lightweight. Since it is defined as the squared $\ell_2$ norm of the mean embedding vector (derived from the continuous hyperspherical density field), its computation only involves global averaging over all node embeddings followed by a vector norm, yielding a complexity of $\mathcal{O}(Nd)$. This avoids the $\mathcal{O}(N^2)$ pairwise comparisons commonly required by many contrastive objectives. Furthermore, the Endogenous Bayesian Thermostat incurs only an $\mathcal{O}(1)$ negligible overhead to update the precision scales ($s_{\mathrm{align}}$ and $s_{\mathrm{unif}}$) based on the instantaneous scalar loss values.

Overall, the total time complexity of {\framework} per epoch is $\mathcal{O}((L+1)|\mathcal{E}|d + Nd)$. When the number of layers $L$ is treated as a small constant, the overall complexity scales linearly with the graph size $\mathcal{O}(|\mathcal{V}| + |\mathcal{E}|)$. This demonstrates that {\framework} maintains the strict scalability of sparse full-graph GNN training while incurring only minimal overhead for structural binding and mean-field repulsive regularization.

\section{Experiments}

To empirically validate the effectiveness and superiority of {\framework}, we design comprehensive experiments to answer the following four core research questions (RQs): (1) How does {\framework} compare with state-of-the-art baselines on node classification and link prediction? (2) Can the Bayesian thermostat adaptively balance structural binding and global dispersion without manual temperature tuning? (3)  Does {\framework} learn a better geometric trade-off between local alignment and global uniformity on the hyperspherical representation manifold? (4) Can the proposed framework generalize across GNN architectures and serve as a plug-and-play mechanism for existing graph self-supervised learning methods?
Additional analyses of embedding dimension and encoder depth are provided in Appendix~\ref{app:robustness}.

\begin{table*}[ht]
\centering
\small
\caption{Node classification accuracy on homophilic graphs. Optimal results are shown in bold and suboptimal results are underlined. OOM indicates Out-Of-Memory on 24GB RTX 3090 GPU.}
\label{tab:homophily_results}
\resizebox{\textwidth}{!}{%
\begin{tabular}{lccccccc}
\toprule
Model & Cora & CiteSeer & PubMed & Computers & Photo & CS & Physics \\
\midrule
GAT        & 80.32 $\pm$ 0.17 & 71.02 $\pm$ 0.28 & 78.83 $\pm$ 0.71 & 86.93 $\pm$ 0.29 & 92.56 $\pm$ 0.35 & 92.31 $\pm$ 0.24 & 95.47 $\pm$ 0.15 \\       
GCN        & 81.54 $\pm$ 0.68 & 70.73 $\pm$ 0.65 & 79.16 $\pm$ 0.25 & 86.51 $\pm$ 0.54 & 92.42 $\pm$ 0.22 & 93.03 $\pm$ 0.31 & 95.65 $\pm$ 0.16 \\
\midrule
DeepWalk   & 70.72 $\pm$ 0.63 & 51.39 $\pm$ 0.41 & 73.27 $\pm$ 0.86 & 85.68 $\pm$ 0.07 & 89.40 $\pm$ 0.11 & 84.61 $\pm$ 0.22 & 91.77 $\pm$ 0.15 \\
Node2Vec   & 71.08 $\pm$ 0.91 & 47.34 $\pm$ 0.84 & 66.23 $\pm$ 0.95 & 84.41 $\pm$ 0.14 & 89.68 $\pm$ 0.19 & 85.16 $\pm$ 0.04 & 91.23 $\pm$ 0.07 \\
\midrule
DGI        & 84.46 $\pm$ 0.09 & 73.30 $\pm$ 0.81 & 85.86 $\pm$ 0.15 & 88.87 $\pm$ 0.20 & 92.83 $\pm$ 0.50 & 93.26 $\pm$ 0.09 & 95.65 $\pm$ 0.11 \\
GRACE      & 83.00 $\pm$ 0.31 & 71.80 $\pm$ 0.92 & 82.87 $\pm$ 0.21 & 85.44 $\pm$ 0.23 & 91.79 $\pm$ 0.17 & 92.65 $\pm$ 0.08 & OOM \\
GCA        & 82.41 $\pm$ 0.55 & 71.56 $\pm$ 0.19 & 80.73 $\pm$ 0.23 & 87.92 $\pm$ 0.33 & 92.35 $\pm$ 0.53 & 92.65 $\pm$ 0.32 & 95.52 $\pm$ 0.21 \\
BGRL       & 80.63 $\pm$ 0.93 & 71.40 $\pm$ 0.72 & 86.00 $\pm$ 0.07 & 89.31 $\pm$ 0.30 & 92.18 $\pm$ 0.12 & 93.30 $\pm$ 0.05 & 94.92 $\pm$ 0.03 \\
SGCL       & 83.17 $\pm$ 0.16 & 72.50 $\pm$ 0.82 & 82.98 $\pm$ 0.31 & \textbf{90.70 $\pm$ 0.30} & 93.46 $\pm$ 0.30 & 93.29 $\pm$ 0.17 & 95.78 $\pm$ 0.11 \\
SGRL       & 81.45 $\pm$ 0.55 & 72.04 $\pm$ 0.54 & 86.24 $\pm$ 0.07 & 89.65 $\pm$ 0.05 & 93.13 $\pm$ 0.07 & 93.60 $\pm$ 0.08 & \underline{96.06 $\pm$ 0.06} \\
Str-GCL    & \underline{85.13 $\pm$ 0.07} & 70.79 $\pm$ 0.85 & \textbf{86.70 $\pm$ 0.27} & \underline{89.80 $\pm$ 0.35} & \underline{93.55 $\pm$ 0.23} & \underline{93.73 $\pm$ 0.12} & OOM \\
GRASS      & 84.93 $\pm$ 1.63 & \underline{73.47 $\pm$ 0.94} & 85.18 $\pm$ 0.76 & 89.76 $\pm$ 0.61 & 93.26 $\pm$ 0.57 & 93.14 $\pm$ 0.03 & OOM \\
\midrule
\framework & \textbf{85.41 $\pm$ 0.11} & \textbf{74.42 $\pm$ 0.23} & \underline{86.29 $\pm$ 0.09} & 88.18 $\pm$ 0.16 & \textbf{93.69 $\pm$ 0.02} & \textbf{94.05 $\pm$ 0.04} & \textbf{96.13 $\pm$ 0.11} \\
\bottomrule
\end{tabular}
}
\end{table*}

\subsection{Experimental Setup}

\subsubsection{Datasets}
\label{sec:dataset}
We evaluate {\framework} on eleven benchmarks. Homophilic graphs include the citation networks Cora, CiteSeer, and PubMed~\citep{kipf2017semi}, Amazon co-purchase networks Computers and Photo, and collaboration networks Coauthor-CS and Coauthor-Physics~\citep{shchur2018pitfalls}. Heterophilic graphs include Actor and the WebKB networks Cornell, Texas, and Wisconsin~\citep{pei2020geomgcn}. Dataset statistics are provided in Appendix~\ref{app:datasets}.

\subsubsection{Baselines}
\label{sec:baselines}

To systematically evaluate {\framework}, we compare it against a diverse set of baselines for node classification and link prediction. For node classification, we include: (1) supervised learning methods, GAT~\citep{velivckovic2018graph} and GCN~\citep{kipf2017semi}; (2) classical graph embedding methods, DeepWalk~\citep{perozzi2014deepwalk} and Node2Vec~\citep{grover2016node2vec}; and (3) GCL methods, DGI~\citep{velickovic2019deep}, GRACE~\citep{zhu2020deep}, GCA~\citep{zhu2021graph}, BGRL~\citep{thakoor2022large}, SGCL~\citep{sun2024rethinking}, SGRL~\citep{he2024exploitation}, Str-GCL~\citep{he2025str}, and GRASS~\citep{yang2025universal}. For link prediction, the baselines comprise: (1) embedding methods, MF~\citep{menon2011link}, MLP, and Node2Vec~\citep{grover2016node2vec}; (2) GNN-based methods, GCN~\citep{kipf2017semi}, GAT~\citep{velivckovic2018graph}, SAGE~\citep{hamilton2017inductive}, and VGAE~\citep{kipf2016variational}; and (3) advanced GNNs for link prediction, SEAL~\citep{zhang2018link}, BUDDY~\citep{chamberlain2023buddy}, NBFNet~\citep{zhu2021neural}, Neo-GNN~\citep{yun2021neo}, PEG~\citep{wang2022peg}, and NCN/NCNC~\citep{wang2024neural}.

\subsubsection{Evaluation and Implementation}
\label{sec:evaluation}
We evaluate node classification using accuracy (\%), and link prediction using AUC-ROC. All results are averaged over five independent runs with different random splits and model initializations. Detailed evaluation protocols and implementation settings are provided in Appendices~\ref{app:evaluation} and~\ref{app:implementation}.

\subsection{Overall Effectiveness (RQ1)}
\label{sec:rq1}

To evaluate {\framework} against state-of-the-art baselines, we conduct node classification and link prediction experiments across diverse graph domains, covering both homophilic and heterophilic structures.

\subsubsection{Node Classification on Homophilic Graphs}

Table~\ref{tab:homophily_results} summarizes the node classification results on seven homophilic graph datasets. {\framework} achieves the best results on Cora, CiteSeer, Amazon-Photo, Coauthor-CS, and Coauthor-Physics, and ranks second on PubMed. It consistently outperforms supervised GCN/GAT and graph embedding methods, while remaining competitive with recent GCL baselines across citation, co-purchase, and collaboration networks. These results suggest that coordinating structural binding and global dispersion produces discriminative representations across diverse homophilic graphs. 
On Amazon-Computers, {\framework} achieves $88.18\%$, below SGCL's $90.70\%$, but maintains strong overall performance across the benchmark suite.

\subsubsection{Node Classification on Heterophilic Graphs}

Table~\ref{tab:heterophily_results} presents the node classification results on heterophilic graphs. Heterophilic graphs are particularly challenging because connected nodes often belong to different classes, making the standard homophily assumption of message-passing GNNs less reliable. As shown in the table, traditional GNNs such as GCN and GAT are generally less competitive on these datasets, suggesting that conventional neighborhood aggregation may introduce noisy or less informative signals when adjacent nodes have different labels. In contrast, {\framework} achieves the best accuracy on Cornell, Texas, and Wisconsin, and ranks second on Actor, demonstrating strong performance across diverse heterophilic structures. These results are consistent with the motivation of our objective: structural binding captures useful neighborhood regularities, while global dispersion counteracts excessive concentration, helping preserve discriminability when connectivity is weakly aligned with class semantics.

\begin{table}[!t]
\centering
\small
\caption{Node classification accuracy on heterophilic graphs.}
\label{tab:heterophily_results}
\setlength{\tabcolsep}{4pt}
\begin{tabular}{lcccc}  
\toprule
Model & Actor & Cornell & Texas & Wisconsin \\
\midrule
GAT        & 28.06 $\pm$ 1.48 & 59.46 $\pm$ 3.63 & 61.62 $\pm$ 3.78 & 54.71 $\pm$ 6.87 \\
GCN        & 30.83 $\pm$ 0.77 & 57.03 $\pm$ 3.30 & 60.00 $\pm$ 4.80 & 56.47 $\pm$ 6.55 \\
\midrule
DeepWalk   & 22.78 $\pm$ 0.64 & 63.35 $\pm$ 4.61 & 60.59 $\pm$ 7.56 & 55.41 $\pm$ 5.96 \\
Node2Vec   & 28.28 $\pm$ 1.27 & 42.94 $\pm$ 7.46 & 41.92 $\pm$ 7.76 & 37.45 $\pm$ 7.09 \\
\midrule
DGI        & 30.81 $\pm$ 1.29 & 67.57 $\pm$ 5.84 & 68.47 $\pm$ 4.59 & 58.63 $\pm$ 3.20 \\
GRACE      & 31.03 $\pm$ 0.74 & 57.66 $\pm$ 8.92 & 58.56 $\pm$ 1.27 & 51.63 $\pm$ 8.06 \\
GCA        & 29.65 $\pm$ 1.47 & 55.41 $\pm$ 4.56 & 59.46 $\pm$ 6.16 & 50.78 $\pm$ 4.06 \\
BGRL       & 28.86 $\pm$ 0.54 & 54.57 $\pm$ 3.06 & 56.92 $\pm$ 3.06 & 51.74 $\pm$ 2.15 \\
SGCL       & 31.23 $\pm$ 0.64 & 62.76 $\pm$ 2.94 & 68.31 $\pm$ 4.33 & 63.26 $\pm$ 3.06 \\
SGRL       & 32.93 $\pm$ 0.12 & 54.35 $\pm$ 4.05 & 61.26 $\pm$ 1.27 & 55.99 $\pm$ 2.02 \\
Str-GCL    & 31.34 $\pm$ 0.87 & 54.05 $\pm$ 4.41 & 63.96 $\pm$ 5.55 & 54.25 $\pm$ 4.03 \\
GRASS      & \textbf{37.43 $\pm$ 0.14} & \underline{72.15 $\pm$ 1.04} & \underline{81.24 $\pm$ 0.76} & \underline{81.48 $\pm$ 0.62} \\
\midrule
\framework & \underline{36.12 $\pm$ 0.35} & \textbf{73.27 $\pm$ 0.85} & \textbf{83.48 $\pm$ 0.42} & \textbf{84.53 $\pm$ 0.82}  \\
\bottomrule
\end{tabular}
\end{table}

\subsubsection{Link Prediction}

Table~\ref{tab:link_prediction_results} reports the link prediction performance in terms of AUC-ROC.  {\framework} achieves the highest AUC-ROC on Cora and CiteSeer and the best average score of $98.62\%$, exceeding NCNC's $97.90\%$. Notably, {\framework} uses a general-purpose representation learning objective yet outperforms specialized predictors such as SEAL, BUDDY, Neo-GNN, NCN, and NCNC in average performance. 
These results suggest that the learned embeddings retain structural proximity while remaining discriminative, consistent with the proposed balance between local structural coherence and global dispersion. Together with the node classification results, these findings support the effectiveness of {\framework} across different downstream tasks.

\begin{table}[!t]
\centering
\small
\caption{Performance on link prediction (AUC-ROC).}
\label{tab:link_prediction_results}
\setlength{\tabcolsep}{4pt}
\begin{tabular}{lcccc}
\toprule
Model & Cora & CiteSeer & PubMed & Avg. \\
\midrule
Node2Vec       & 90.97 $\pm$ 0.64 & 94.46 $\pm$ 0.59 & 93.14 $\pm$ 0.18 & 92.86 \\
MF             & 80.29 $\pm$ 2.26 & 75.92 $\pm$ 3.25 & 93.06 $\pm$ 0.43 & 83.09 \\
MLP            & 95.32 $\pm$ 0.37 & 94.45 $\pm$ 0.32 & 98.34 $\pm$ 0.10 & 96.04 \\
\midrule
GCN            & 95.01 $\pm$ 0.32 & 95.89 $\pm$ 0.26 & 98.69 $\pm$ 0.06 & 96.53 \\
GAT            & 93.90 $\pm$ 0.32 & 96.25 $\pm$ 0.20 & 98.20 $\pm$ 0.07 & 96.12 \\
SAGE           & 95.63 $\pm$ 0.27 & 97.39 $\pm$ 0.15 & 98.87 $\pm$ 0.04 & 97.30 \\
VGAE           & 95.08 $\pm$ 0.33 & 97.06 $\pm$ 0.22 & 97.47 $\pm$ 0.08 & 96.54 \\
\midrule
SEAL           & 90.59 $\pm$ 0.75 & 88.52 $\pm$ 1.40 & 97.77 $\pm$ 0.40 & 92.29 \\
BUDDY          & 95.06 $\pm$ 0.36 & 96.72 $\pm$ 0.26 & 98.20 $\pm$ 0.05 & 96.66 \\
Neo-GNN        & 93.73 $\pm$ 0.36 & 94.89 $\pm$ 0.60 & 98.71 $\pm$ 0.05 & 95.78 \\
NCN            & 96.76 $\pm$ 0.18 & 97.04 $\pm$ 0.26 & \underline{98.98 $\pm$ 0.04} & 97.59 \\
NCNC           & \underline{96.90 $\pm$ 0.28} & \underline{97.65 $\pm$ 0.30} & \textbf{99.14 $\pm$ 0.03} & \underline{97.90} \\
NBFNet         & 92.85 $\pm$ 0.17 & 91.06 $\pm$ 0.15 & 98.34 $\pm$ 0.02 & 94.08 \\
PEG            & 94.46 $\pm$ 0.34 & 96.15 $\pm$ 0.41 & 96.97 $\pm$ 0.39 & 95.86 \\
\midrule
\framework     & \textbf{98.10 $\pm$ 0.40} & \textbf{99.01 $\pm$ 0.14} & 98.75 $\pm$ 0.08 & \textbf{98.62} \\
\bottomrule
\end{tabular}
\end{table}

\subsection{Adaptive Thermostat Dynamics (RQ2)}
\label{sec:rq2}

\begin{figure}[!t]
    \centering
    \subfloat[Cora]{%
        \includegraphics[width=0.49\columnwidth]{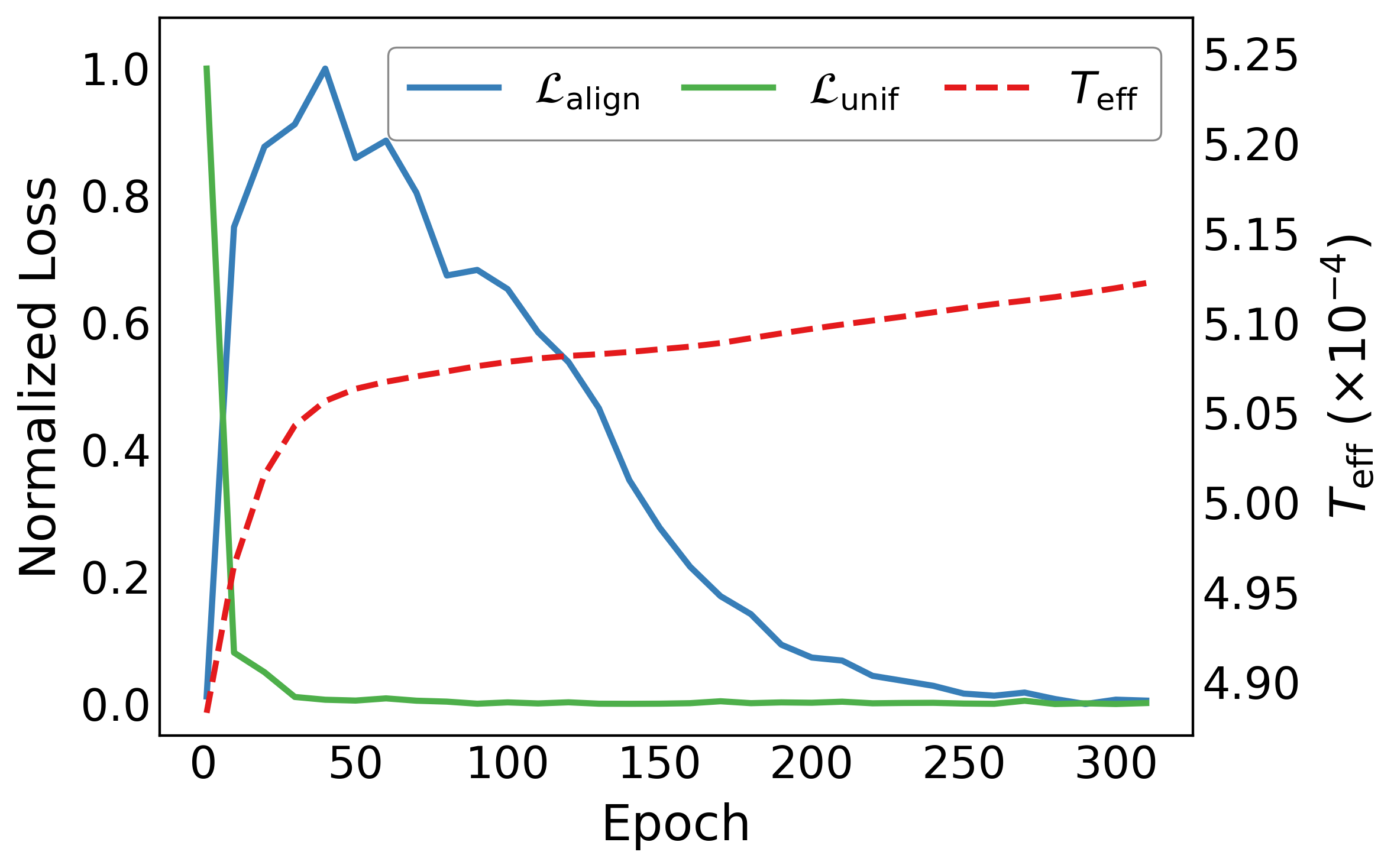}
    }
    \subfloat[Texas]{%
        \includegraphics[width=0.49\columnwidth]{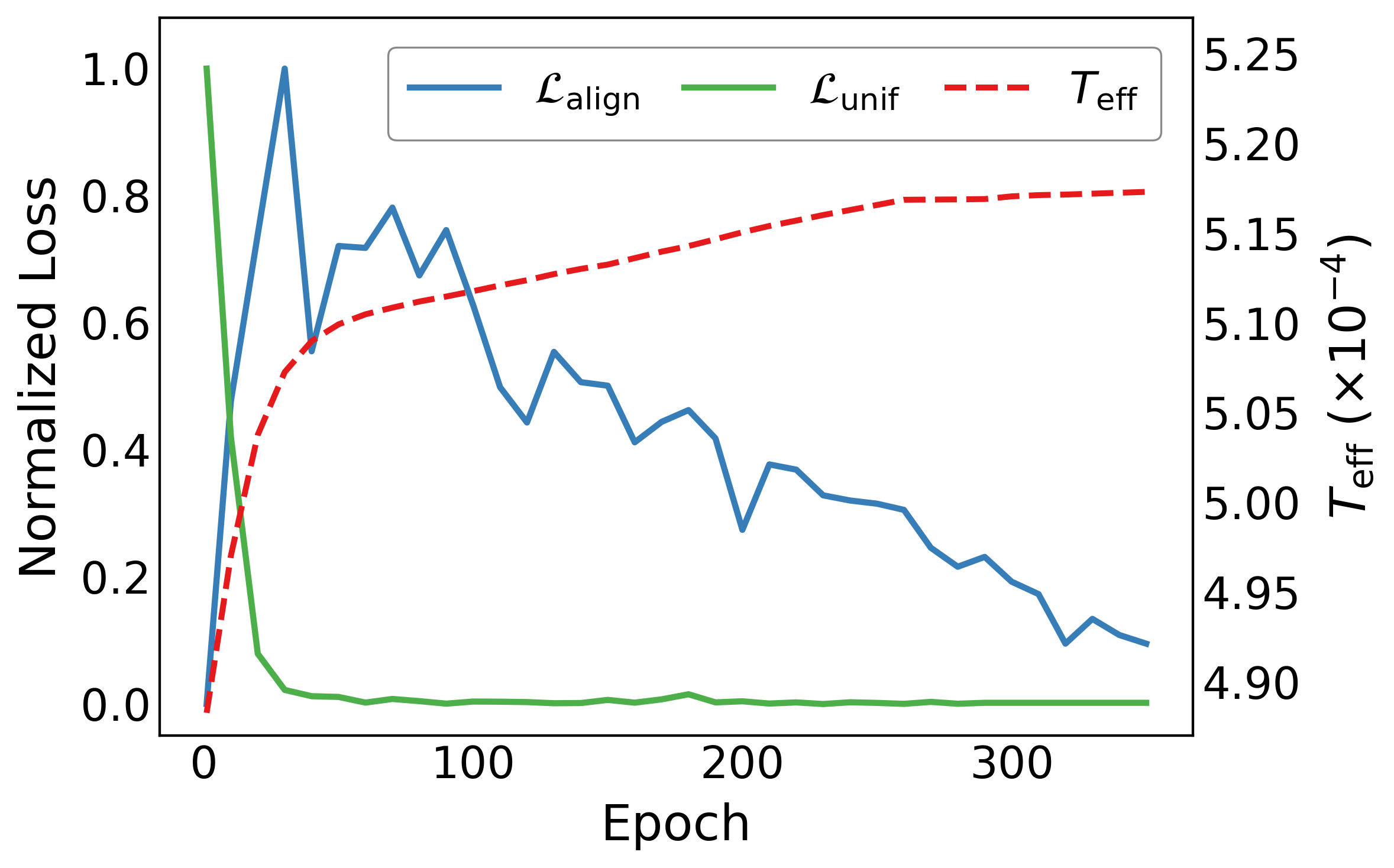}
    }
    \caption{Thermodynamic evolution on Cora and Texas, showing the training trajectories of the normalized losses $\mathcal{L}_{\mathrm{align}}$ and $\mathcal{L}_{\mathrm{unif}}$, together with the effective temperature $T_{\mathrm{eff}}$.}
    \label{fig:thermostat_dynamics}
\end{figure}


\begin{figure}[!t]
    \centering
    \includegraphics[width=0.5\columnwidth]{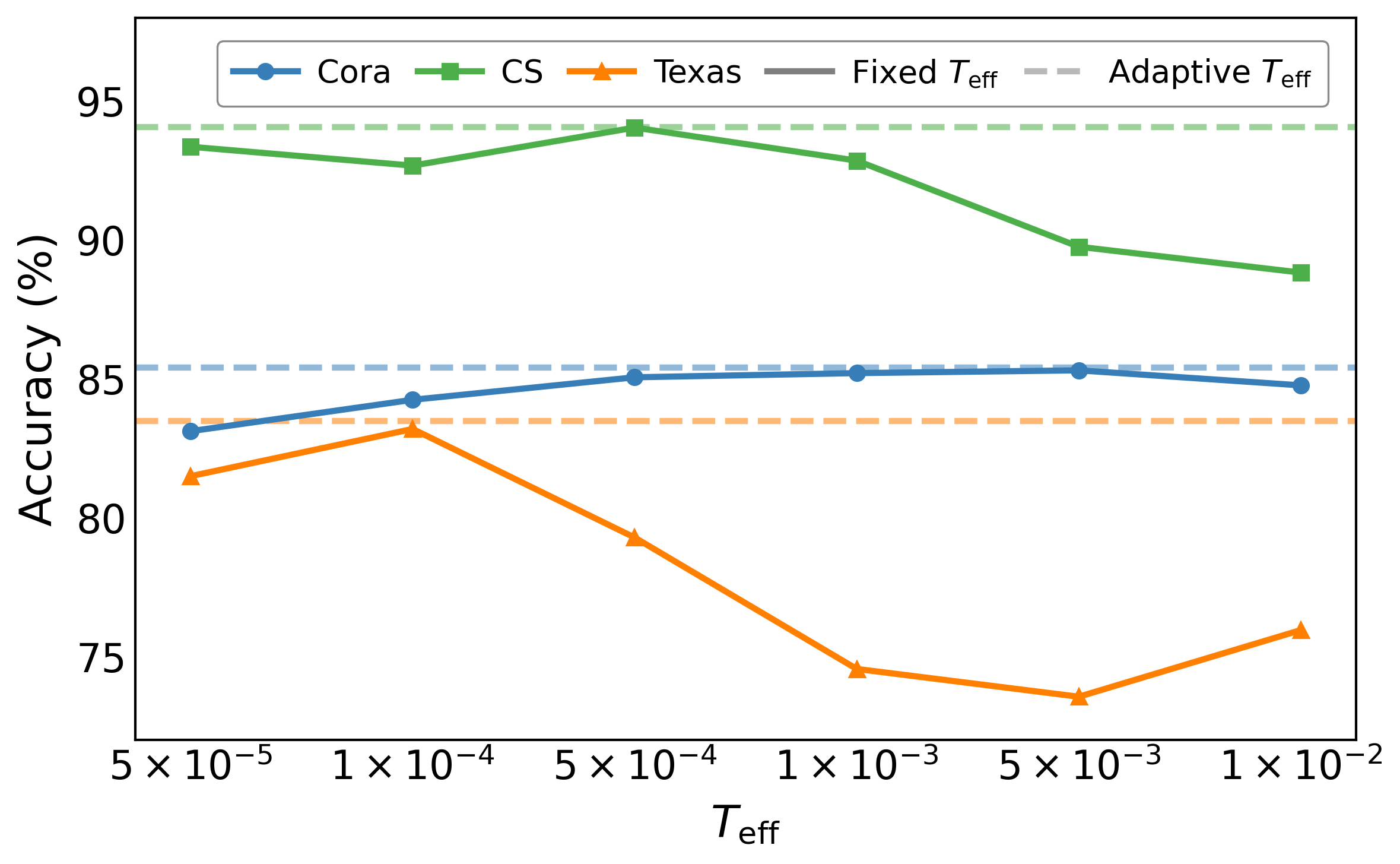}
    \caption{Sensitivity analysis of the fixed effective temperature on Cora, CS, and Texas.}
    \label{fig:fixed_temperature_sensitivity}
\end{figure}

Fig.~\ref{fig:thermostat_dynamics} shows the training dynamics on Cora and Texas, with independently normalized losses and $T_{\mathrm{eff}}$ on its original scale. On both datasets, $\mathcal{L}_{\mathrm{unif}}$ rapidly decreases and remains near zero, indicating reduced global concentration on the hypersphere, while $\mathcal{L}_{\mathrm{align}}$ initially rises before declining, suggesting that early redistribution is followed by improved local structural coherence. Meanwhile, $T_{\mathrm{eff}}$ continues to increase on Cora but approaches a plateau on Texas. These dataset-dependent dynamics indicate that the thermostat responds adaptively to the evolving representation geometry instead of imposing a rigid, uniform temperature schedule across diverse graph structures.

Fig.~\ref{fig:fixed_temperature_sensitivity} compares adaptive and fixed temperatures on Cora, CS, and Texas. The preferred fixed temperature varies across datasets, with Texas showing particularly pronounced performance degradation at larger values. 
In contrast, the adaptive thermostat achieves consistently competitive performance without selecting a dataset-specific temperature. 
Together, these findings support regulating structural binding and global dispersion dynamically according to the evolving representation state.

\subsection{Representation Geometry (RQ3)}
\label{sec:rq3}

Fig.~\ref{fig:tsne_cora} in Appendix~\ref{app:representation_visualization} visualizes Cora embeddings using t-SNE~\citep{maaten2008visualizing}, with colors indicating node classes. Raw features exhibit substantial inter-class overlap, while GRACE, DGI, BGRL, and SGRL reveal clearer clusters but retain fragmented groups and mixed regions. In comparison, {\framework} produces more distinct class groups with reduced mixing in several regions, although some overlap and fragmentation remain.
These observations qualitatively support a favorable balance between intra-class compactness and inter-class separation.

\begin{figure}[!t]
    \centering
    \subfloat[Cora]{%
        \includegraphics[width=0.49\columnwidth]{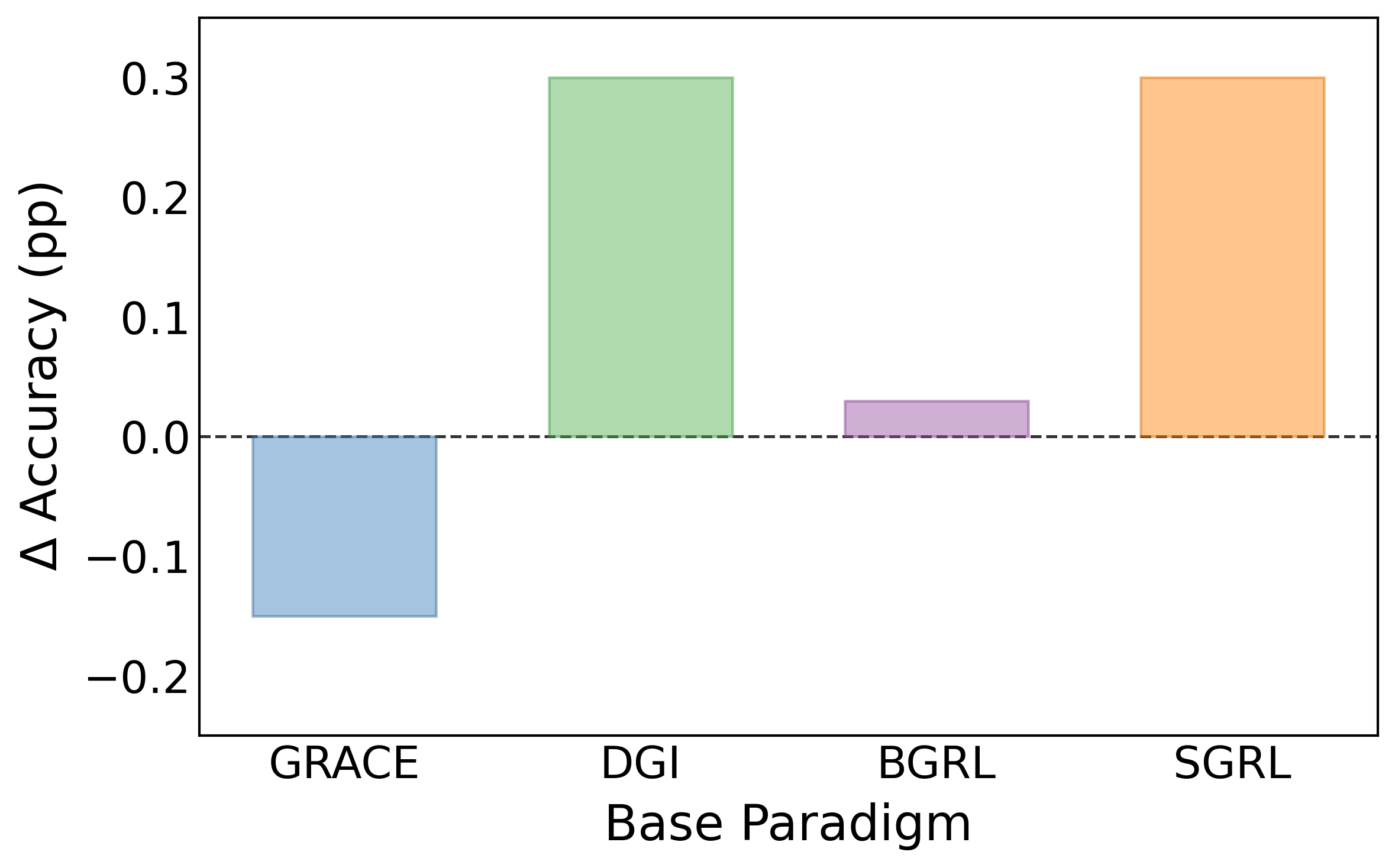}
    }\hfill
    \subfloat[Texas]{%
        \includegraphics[width=0.49\columnwidth]{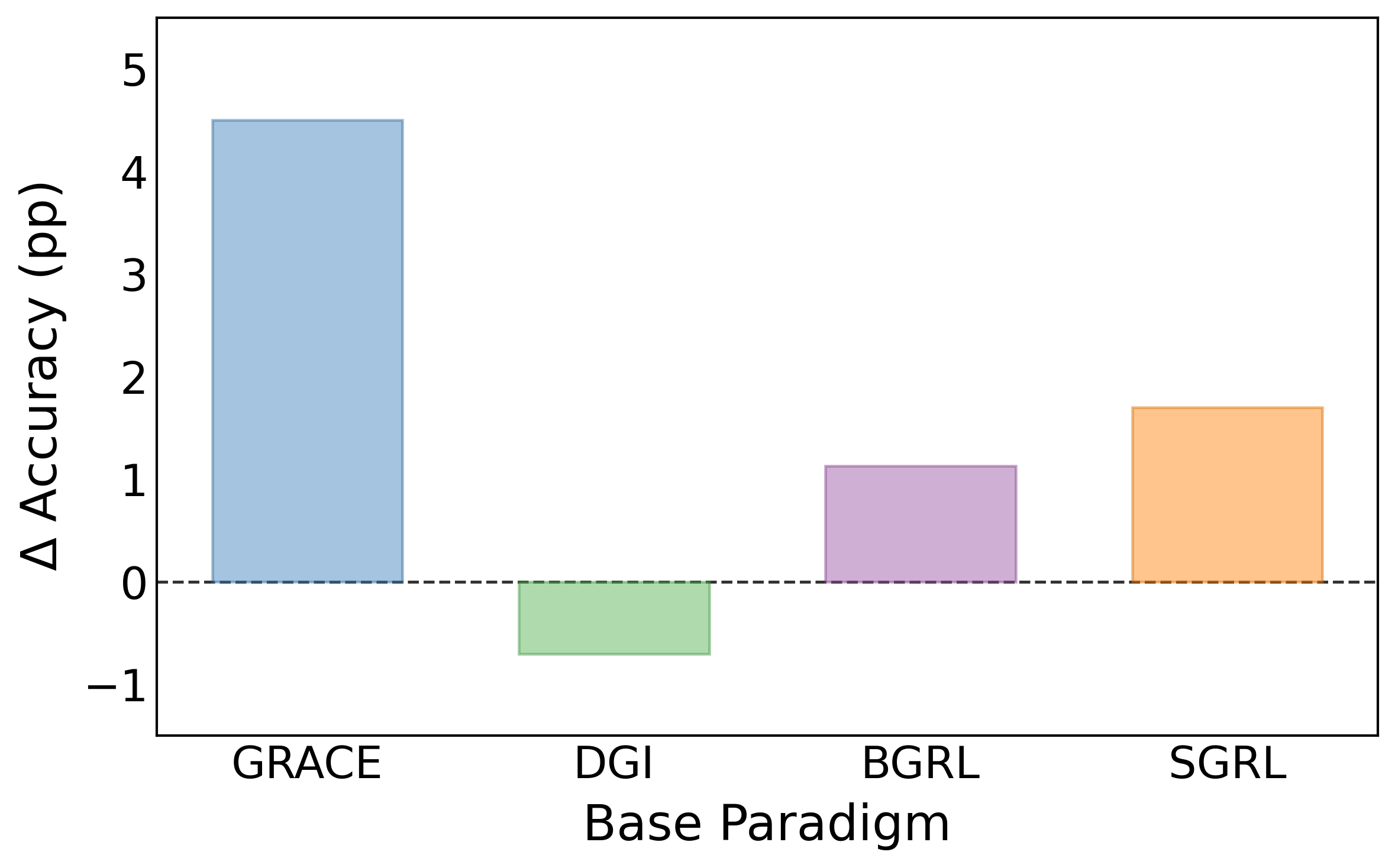}
    }
    \caption{Accuracy changes (percentage points) with Bayesian thermostat integration.}
    \label{fig:paradigm_universality}
\end{figure}

\subsection{Paradigm Universality (RQ4)}
\label{sec:rq4}

\subsubsection{Generality across GNN Backbones.}
Table~\ref{tab:backbone} evaluates {\framework} with different backbones. On homophilic Cora and CS datasets, neighboring nodes tend to share class labels, making neighborhood aggregation informative across architectures and yielding similar performance. On the heterophilic Texas dataset, edges frequently connect nodes of different classes, causing neighborhood mixing to introduce semantic noise that severely degrades GCN and GAT. In contrast, GraphSAGE and Transformer achieve substantially higher accuracy; their capacity to decouple or selectively emphasize central-node features helps preserve node-specific semantics against misleading neighborhood aggregation.
Overall, these results confirm the backbone compatibility of {\framework} while highlighting the necessity of preserving node-specific information under heterophily.

\begin{table}[ht]
\centering
\footnotesize
\caption{Accuracy (\%) with different GNN backbones.}
\label{tab:backbone}
\begin{tabular}{lccc}
\toprule
Backbone & Cora & CS & Texas \\
\midrule
GCN              & 84.78 $\pm$ 0.49 & 93.57 $\pm$ 0.03 & 48.95 $\pm$ 1.12 \\
GAT              & 85.10 $\pm$ 0.23 & 93.42 $\pm$ 0.08 & 58.56 $\pm$ 0.74 \\
GraphSAGE        & 84.87 $\pm$ 0.44 & 93.71 $\pm$ 0.05 & 79.19 $\pm$ 0.74 \\
Transformer      & \textbf{85.41 $\pm$ 0.11} & \textbf{94.05 $\pm$ 0.04} & \textbf{83.48 $\pm$ 0.42} \\
\bottomrule
\end{tabular}
\end{table}

\subsubsection{Plug-and-Play Integration}

We apply the Bayesian thermostat to GRACE, DGI, BGRL, and SGRL to examine its transferability across graph self-supervised objectives. For GRACE, DGI, and SGRL, we retain their objective-specific attractive and dispersive components, express them as non-negative losses, and adaptively balance them using the thermostat. For BGRL, which lacks an explicit dispersion objective, we retain its prediction loss and introduce the sampling-free dispersion term as a complementary component.

Fig.~\ref{fig:paradigm_universality} shows improvements in most method--dataset combinations on Cora and Texas, with larger gains for GRACE, BGRL, and SGRL on Texas. GRACE declines slightly on Cora, while DGI declines on Texas. These results suggest that the thermostat can complement diverse objectives, although its benefits depend on the method and graph structure.

Table~\ref{tab:bgrl_uniformity} further distinguishes the effect of adding a uniformity term with adaptively balancing prediction and dispersion. BGRL-FU adds a fixed-weight uniformity term whose coefficient is selected on validation data, but yields no improvement over BGRL on the evaluated datasets. BGRL-BT instead achieves slightly higher mean accuracy on Cora and CS and a larger gain on Texas. 
These results suggest that the benefit does not arise solely from introducing an additional uniformity term; training-adaptive coordination through the Bayesian thermostat provides a more favorable balance between prediction and dispersion.

\begin{table}[t]
\centering
\footnotesize
\caption{Node classification accuracy (\%) of BGRL variants. FU denotes fixed-weight uniformity, and BT denotes the Bayesian thermostat.}
\label{tab:bgrl_uniformity}
\begin{tabular}{lccc}
\toprule
Method & Cora & CS & Texas \\
\midrule
BGRL       & 80.63 $\pm$ 0.93 & 93.30 $\pm$ 0.05 & 56.92 $\pm$ 3.06 \\
BGRL-FU    & 80.48 $\pm$ 1.11 & 93.28 $\pm$ 0.02 & 56.72 $\pm$ 3.06 \\
BGRL-BT    & 80.66 $\pm$ 0.76 & 93.35 $\pm$ 0.14 & 58.05 $\pm$ 4.46 \\
\bottomrule
\end{tabular}
\end{table}

\section{Related Work}

\textbf{Graph Neural Networks.} Most Graph Neural Networks construct node representations through recursive neighborhood aggregation, including GCN~\citep{kipf2017semi}, GraphSAGE~\citep{hamilton2017inductive}, GAT~\citep{velivckovic2018graph}, and GIN~\citep{xu2019how}, while graph Transformers combine message passing with flexible attention~\citep{rampavsek2022recipe}. Repeated propagation promotes local consistency but can cause over-smoothing and reduce discriminability~\citep{li2018deeper,oono2020graph}. Normalization, topological regularization, and modified propagation mitigate this issue~\citep{zhao2020pairnorm,chen2020measuring,chen2020simple}, yet maintaining globally distinguishable representations remains challenging.

\textbf{Graph Contrastive Learning.} Graph Contrastive Learning complements neighborhood aggregation with self-supervised objectives that preserve semantic consistency and representation diversity. GRACE~\citep{zhu2020deep} and GCA~\citep{zhu2021graph} contrast nodes across augmented views; DGI~\citep{velickovic2019deep} and MVGRL~\citep{hassani2020contrastive} contrast local--global pairs against corrupted samples; and BGRL~\citep{thakoor2022large} and CCA-SSG~\citep{zhang2021canonical} avoid explicit negative pairs through bootstrapping and feature decorrelation, respectively. Their alignment--dispersion balance depends on method-specific hyperparameters, augmentations, or architectural mechanisms, motivating explicit regulation based on evolving representation geometry.

\textbf{Hyperspherical Alignment and Uniformity.}
The representation-diversity mechanisms in GCL can be viewed through the geometric perspective of alignment and uniformity~\citep{wang2020understanding}. 
Alignment encourages semantically related representations to remain close, while uniformity promotes global dispersion on the unit hypersphere~\citep{wang2020understanding}. HCAN~\citep{fang2021hyperspherical} and DAGC~\citep{wang2024deep} model hyperspherical representations with directional distributions; HPNC~\citep{lu2024hyperspherical} improves inter-cluster separation through hyperspherical prototypes; and SGRL~\citep{he2024exploitation} promotes representation scattering without conventional negative sampling. 
These approaches primarily emphasize distribution-level uniformity, prototype separation, or representation scattering, with local graph structure often incorporated indirectly through encoders or augmentations. 
Thus, local cohesion and hyperspherical dispersion are often modeled separately and balanced by predefined regularizers or objective weights. A unified formulation that directly couples structure-aware neighborhood binding with global hyperspherical entropy, allowing a balanced representation geometry to emerge through joint optimization, remains largely underexplored.

\section{Conclusion} 

In this paper, we introduced {\framework}, a self-organized framework that bridges statistical thermodynamics and graph self-supervised learning. Rather than relying on heuristic objective designs, our approach establishes a principled paradigm by minimizing the Helmholtz free energy on the unit hypersphere. We demonstrated that the tension between local structural cohesion and global dispersion can be resolved through a unified thermodynamic objective, explicitly balancing structural binding energy and representational entropy. Furthermore, the integration of an Endogenous Bayesian Thermostat mitigates the need for manually specified loss weights by enabling the system to autonomously adjust toward a metastable semantic equilibrium.
Experiments across diverse graphs verify the effectiveness and robustness of the representations in node classification and link prediction. Despite these advancements, future work will explore heterophily-aware structural anchors and scalable adaptive extensions for large and dynamic graphs.

\bibliographystyle{ACM-Reference-Format}
\bibliography{references}

@inproceedings{zhu2021graph,
  title={Graph contrastive learning with adaptive augmentation},
  author={Zhu, Yanqiao and Xu, Yichen and Yu, Feng and Liu, Qiang and Wu, Shu and Wang, Liang},
  booktitle={Proceedings of the ACM Web Conference},
  pages={2069--2080},
  year={2021}
}

@article{he2024exploitation,
  title={Exploitation of a latent mechanism in graph contrastive learning: Representation scattering},
  author={He, Dongxiao and Shan, Lianze and Zhao, Jitao and Zhang, Hengrui and Wang, Zhen and Zhang, Weixiong},
  journal={Advances in Neural Information Processing Systems},
  volume={37},
  pages={115351--115376},
  year={2024}
}

@inproceedings{thakoor2022large,
  title={Large-Scale Representation Learning on Graphs via Bootstrapping},
  author={Thakoor, Shantanu and Tallec, Corentin and Azar, Mohammad Gheshlaghi and Azabou, Mehdi and Dyer, Eva L and Munos, Remi and Veli{\v{c}}kovi{\'c}, Petar and Valko, Michal},
  booktitle={Proceedings of the International Conference on Learning Representations},
  year={2022}
}

@article{zhang2021canonical,
  title={From canonical correlation analysis to self-supervised graph neural networks},
  author={Zhang, Hengrui and Wu, Qitian and Yan, Junchi and Wipf, David and Yu, Philip S},
  journal={Advances in Neural Information Processing Systems},
  volume={34},
  pages={76--89},
  year={2021}
}

@misc{kipf2016variational,
  title={Variational Graph Auto-Encoders},
  author={Kipf, Thomas N and Welling, Max},
  note={arXiv preprint arXiv:1611.07308},
  year={2016}
}

@inproceedings{hassani2020contrastive,
  title={Contrastive multi-view representation learning on graphs},
  author={Hassani, Kaveh and Khasahmadi, Amir Hosein},
  booktitle={Proceedings of the International Conference on Machine Learning},
  pages={4116--4126},
  year={2020},
  organization={PMLR}
}

@inproceedings{velickovic2019deep,
  title={Deep Graph Infomax},
  author={Veli{\v{c}}kovi{\'c}, Petar and Fedus, William and Hamilton, William L and Li{\`o}, Pietro and Bengio, Yoshua and Hjelm, R Devon},
  booktitle={Proceedings of the International Conference on Learning Representations},
  year={2019}
}

@misc{zhu2020deep,
  title={Deep graph contrastive representation learning},
  author={Zhu, Yanqiao and Xu, Yichen and Yu, Feng and Liu, Qiang and Wu, Shu and Wang, Liang},
  note={arXiv preprint arXiv:2006.04131},
  year={2020}
}

@article{wang2024deep,
  title={Deep adaptive graph clustering via von Mises-Fisher distributions},
  author={Wang, Pengfei and Wu, Daqing and Chen, Chong and Liu, Kunpeng and Fu, Yanjie and Huang, Jianqiang and Zhou, Yuanchun and Zhan, Jianfeng and Hua, Xiansheng},
  journal={ACM Transactions on the Web},
  volume={18},
  number={2},
  pages={1--21},
  year={2024},
  publisher={ACM New York, NY}
}

@inproceedings{wang2020understanding,
  title={Understanding contrastive representation learning through alignment and uniformity on the hypersphere},
  author={Wang, Tongzhou and Isola, Phillip},
  booktitle={Proceedings of the International Conference on Machine Learning},
  pages={9929--9939},
  year={2020},
  organization={PMLR}
}

@misc{shchur2018pitfalls,
  title={Pitfalls of graph neural network evaluation},
  author={Shchur, Oleksandr and Mumme, Maximilian and Bojchevski, Aleksandar and G{\"u}nnemann, Stephan},
  note={arXiv preprint arXiv:1811.05868},
  year={2018}
}

@inproceedings{kipf2017semi,
  title     = {Semi-Supervised Classification with Graph Convolutional Networks},
  author    = {Kipf, Thomas N. and Welling, Max},
  booktitle = {Proceedings of the International Conference on Learning Representations},
  year      = {2017}
}

@inproceedings{perozzi2014deepwalk,
  title={Deepwalk: Online learning of social representations},
  author={Perozzi, Bryan and Al-Rfou, Rami and Skiena, Steven},
  booktitle={Proceedings of the ACM SIGKDD conference on Knowledge Discovery and Data Mining},
  pages={701--710},
  year={2014}
}

@inproceedings{grover2016node2vec,
  title={Node2Vec: Scalable feature learning for networks},
  author={Grover, Aditya and Leskovec, Jure},
  booktitle={Proceedings of the ACM SIGKDD conference on Knowledge Discovery and Data Mining},
  pages={855--864},
  year={2016}
}

@inproceedings{sun2024rethinking,
  title={Rethinking and simplifying bootstrapped graph latents},
  author={Sun, Wangbin and Li, Jintang and Chen, Liang and Wu, Bingzhe and Bian, Yatao and Zheng, Zibin},
  booktitle={Proceedings of the ACM International Conference on Web Search and Data Mining},
  pages={665--673},
  year={2024}
}

@article{maaten2008visualizing,
  title={Visualizing data using t-SNE},
  author={Maaten, Laurens van der and Hinton, Geoffrey},
  journal={Journal of Machine Learning Research},
  volume={9},
  pages={2579--2605},
  year={2008}
}

@article{lu2024hyperspherical,
  title={Hyperspherical Prototype Node Clustering},
  author={Lu, Jitao and Wu, Danyang and Nie, Feiping and Wang, Rong and Li, Xuelong},
  journal={Transactions on Machine Learning Research},
  pages={1--18},
  year={2024}
}

@article{fang2021hyperspherical,
  title={Hyperspherical variational co-embedding for attributed networks},
  author={Fang, Jinyuan and Liang, Shangsong and Meng, Zaiqiao and De Rijke, Maarten},
  journal={ACM Transactions on Information Systems},
  volume={40},
  number={3},
  pages={1--36},
  year={2021},
  publisher={ACM New York, NY}
}

@article{li2023evaluating,
  title={Evaluating graph neural networks for link prediction: Current pitfalls and new benchmarking},
  author={Li, Juanhui and Shomer, Harry and Mao, Haitao and Zeng, Shenglai and Ma, Yao and Shah, Neil and Tang, Jiliang and Yin, Dawei},
  journal={Advances in Neural Information Processing Systems},
  volume={36},
  pages={3853--3866},
  year={2023}
}

@inproceedings{menon2011link,
  title={Link prediction via matrix factorization},
  author={Menon, Aditya Krishna and Elkan, Charles},
  booktitle={Joint European Conference on Machine Learning and Knowledge Discovery in Databases},
  pages={437--452},
  year={2011},
  organization={Springer}
}

@inproceedings{velivckovic2018graph,
  title={Graph attention networks},
  author={Veli{\v{c}}kovi{\'c}, Petar and Cucurull, Guillem and Casanova, Arantxa and Romero, Adriana and Lio, Pietro and Bengio, Yoshua},
  booktitle={Proceedings of the International Conference on Learning Representations},
  year={2018}
}

@article{hamilton2017inductive,
  title={Inductive representation learning on large graphs},
  author={Hamilton, Will and Ying, Zhitao and Leskovec, Jure},
  journal={Advances in Neural Information Processing Systems},
  volume={30},
  year={2017}
}

@article{zhang2018link,
  title={Link prediction based on graph neural networks},
  author={Zhang, Muhan and Chen, Yixin},
  journal={Advances in Neural Information Processing Systems},
  volume={31},
  year={2018}
}

@inproceedings{chamberlain2023buddy,
  title={Graph Neural Networks for Link Prediction with Subgraph Sketching},
  author={Chamberlain, Benjamin Paul and Shirobokov, Sergey and Rossi, Emanuele and Frasca, Fabrizio and Markovich, Thomas and Hammerla, Nils Yannick and Bronstein, Michael M. and Hansmire, Max},
  booktitle={Proceedings of the International Conference on Learning Representations},
  year={2023}
}

@article{zhu2021neural,
  title={Neural bellman-ford networks: A general graph neural network framework for link prediction},
  author={Zhu, Zhaocheng and Zhang, Zuobai and Xhonneux, Louis-Pascal and Tang, Jian},
  journal={Advances in Neural Information Processing Systems},
  volume={34},
  pages={29476--29490},
  year={2021}
}

@article{yun2021neo,
  title={Neo-gnns: Neighborhood overlap-aware graph neural networks for link prediction},
  author={Yun, Seongjun and Kim, Seoyoon and Lee, Junhyun and Kang, Jaewoo and Kim, Hyunwoo J},
  journal={Advances in Neural Information Processing Systems},
  volume={34},
  pages={13683--13694},
  year={2021}
}

@inproceedings{
  wang2024neural,
  title={Neural Common Neighbor with Completion for Link Prediction},
  author={Xiyuan Wang and Haotong Yang and Muhan Zhang},
  booktitle={Proceedings of the International Conference on Learning Representations},
  year={2024}
}

@inproceedings{wang2022peg,
  title={Equivariant and Stable Positional Encoding for More Powerful Graph Neural Networks},
  author={Wang, Haorui and Yin, Haoteng and Zhang, Muhan and Li, Pan},
  booktitle={Proceedings of the International Conference on Learning Representations},
  year={2022}
}

@misc{oord2018representation,
  title={Representation learning with contrastive predictive coding},
  author={Oord, Aaron van den and Li, Yazhe and Vinyals, Oriol},
  note={arXiv preprint arXiv:1807.03748},
  year={2018}
}

@article{grill2020bootstrap,
  title={Bootstrap your own latent-a new approach to self-supervised learning},
  author={Grill, Jean-Bastien and Strub, Florian and Altch{\'e}, Florent and Tallec, Corentin and Richemond, Pierre and Buchatskaya, Elena and Doersch, Carl and Avila Pires, Bernardo and Guo, Zhaohan and Gheshlaghi Azar, Mohammad and others},
  journal={Advances in Neural Information Processing Systems},
  volume={33},
  pages={21271--21284},
  year={2020}
}

@inproceedings{li2018deeper,
  title={Deeper insights into graph convolutional networks for semi-supervised learning},
  author={Li, Qimai and Han, Zhichao and Wu, Xiao-Ming},
  booktitle={Proceedings of the AAAI conference on Artificial Intelligence},
  volume={32},
  number={1},
  pages={3538--3545},
  year={2018}
}

@inproceedings{oono2020graph,
  title={Graph Neural Networks Exponentially Lose Expressive Power for Node Classification},
  author={Oono, Kenta and Suzuki, Taiji},
  booktitle={Proceedings of the International Conference on Learning Representations},
  year={2020}
}

@article{you2020graphcl,
  title     = {Graph Contrastive Learning with Augmentations},
  author    = {You, Yuning and Chen, Tianlong and Sui, Yongduo and Chen, Ting and Wang, Zhangyang and Shen, Yang},
  journal   = {Advances in Neural Information Processing Systems},
  volume    = {33},
  pages     = {5812--5823},
  year      = {2020}
}

@inproceedings{he2025str,
  title={Str-gcl: Structural commonsense driven graph contrastive learning},
  author={He, Dongxiao and Huang, Yongqi and Zhao, Jitao and Wang, Xiaobao and Wang, Zhen},
  booktitle={Proceedings of the ACM Web Conference},
  pages={1129--1141},
  year={2025}
}

@inproceedings{yang2025universal,
  title={Universal graph self-contrastive learning},
  author={Yang, Liang and Cai, Yukun and Ning, Hui and Zhuo, Jiaming and Jin, Di and Ma, Ziyi and Guo, Yuanfang and Wang, Chuan and Wang, Zhen},
  booktitle={Proceedings of the International Joint Conference on Artificial Intelligence},
  pages={3534--3542},
  year={2025}
}

@article{liu2022graph,
  title={Graph self-supervised learning: A survey},
  author={Liu, Yixin and Jin, Ming and Pan, Shirui and Zhou, Chuan and Zheng, Yu and Xia, Feng and Yu, Philip S},
  journal={IEEE Transactions on Knowledge and Data Engineering},
  volume={35},
  number={6},
  pages={5879--5900},
  year={2022},
  publisher={IEEE}
}

@article{zhang2020deep,
  title={Deep learning on graphs: A survey},
  author={Zhang, Ziwei and Cui, Peng and Zhu, Wenwu},
  journal={IEEE Transactions on Knowledge and Data Engineering},
  volume={34},
  number={1},
  pages={249--270},
  year={2020},
  publisher={IEEE}
}

@inproceedings{zhao2020pairnorm,
  title     = {PairNorm: Tackling Oversmoothing in GNNs},
  author    = {Zhao, Lingxiao and Akoglu, Leman},
  booktitle = {Proceedings of the International Conference on Learning Representations},
  year      = {2020}
}

@inproceedings{pei2020geomgcn,
  title={Geom-GCN: Geometric Graph Convolutional Networks},
  author={Pei, Hongbin and Wei, Bingzhe and Chang, Kevin Chen-Chuan and Lei, Yu and Yang, Bo},
  booktitle={Proceedings of the International Conference on Learning Representations},
  year = {2020}
}

@inproceedings{chen2020measuring,
  title={Measuring and relieving the over-smoothing problem for graph neural networks from the topological view},
  author={Chen, Deli and Lin, Yankai and Li, Wei and Li, Peng and Zhou, Jie and Sun, Xu},
  booktitle={Proceedings of the AAAI conference on Artificial Intelligence},
  volume={34},
  number={04},
  pages={3438--3445},
  year={2020}
}

@inproceedings{chen2020simple,
  title     = {Simple and Deep Graph Convolutional Networks},
  author    = {Chen, Ming and Wei, Zhewei and Huang, Zengfeng and Ding, Bolin and Li, Yaliang},
  booktitle = {Proceedings of the International Conference on Machine Learning},
  pages     = {1725--1735},
  year      = {2020},
  organization={PMLR}
}

@article{ju2024comprehensive,
  title={A comprehensive survey on deep graph representation learning},
  author={Ju, Wei and Fang, Zheng and Gu, Yiyang and Liu, Zequn and Long, Qingqing and Qiao, Ziyue and Qin, Yifang and Shen, Jianhao and Sun, Fang and Xiao, Zhiping and others},
  journal={Neural Networks},
  volume={173},
  pages={106207},
  year={2024},
  publisher={Elsevier}
}

@inproceedings{xu2019how,
  title={How Powerful are Graph Neural Networks?},
  author={Xu, Keyulu and Hu, Weihua and Leskovec, Jure and Jegelka, Stefanie},
  booktitle={Proceedings of the International Conference on Learning Representations},
  year = {2019}
}

@article{rampavsek2022recipe,
  title={Recipe for a general, powerful, scalable graph transformer},
  author={Ramp{\'a}{\v{s}}ek, Ladislav and Galkin, Michael and Dwivedi, Vijay Prakash and Luu, Anh Tuan and Wolf, Guy and Beaini, Dominique},
  journal={Advances in Neural Information Processing Systems},
  volume={35},
  pages={14501--14515},
  year={2022}
}

@book{pathria2011statistical,
  title={Statistical Mechanics},
  author={Pathria, R. K. and Beale, Paul D.},
  year={2011},
  publisher={Academic Press},
}

@inproceedings{kendall2018multi,
  title={Multi-task learning using uncertainty to weigh losses for scene geometry and semantics},
  author={Kendall, Alex and Gal, Yarin and Cipolla, Roberto},
  booktitle={Proceedings of the IEEE Conference on Computer Vision and Pattern Recognition},
  pages={7482--7491},
  year={2018}
}


\appendix
\raggedbottom

\section{Glossary of Notations}
\label{app:notations}
Table~\ref{tab:notations} summarizes the key notation used in this paper.

\begin{table}[!t]
    \centering
    \caption{Summary of key notations.}
    \label{tab:notations}
    \renewcommand{\arraystretch}{1.05}
    \small
    \begin{tabularx}{\columnwidth}{@{}lX@{}}
        \toprule
        Symbol & Description \\
        \midrule
        $\mathcal{G}=(\mathcal{V},\mathcal{E},\mathbf{A},\mathbf{X})$
        & Attributed input graph with node set $\mathcal{V}$, edge set
        $\mathcal{E}$, adjacency matrix $\mathbf{A}$, and node feature
        matrix $\mathbf{X}$. \\

        $N,F,d$
        & Number of nodes, input feature dimension, and embedding dimension. \\

        $f_{\boldsymbol{\theta}}(\cdot)$ 
        & Graph encoder parameterized by $\boldsymbol{\theta}$. \\

        $\mathbf{H},\mathbf{Z}$
        & Intermediate and $\ell_2$-normalized embedding matrices. \\

        $\mathbb{S}^{d-1}$
        & $(d-1)$-dimensional unit hypersphere. \\

        $\mathcal{N}_i,\boldsymbol{\mu}_i$
        & Neighbor set and normalized neighbor-mean structural anchor of
        node $i$. \\

        $p(\mathbf{z}),p^*(\mathbf{z})$
        & Learned representation density and the corresponding
        Boltzmann--Gibbs equilibrium density. \\

        $\hat{p}_{\kappa}(\mathbf{z})$
        & vMF-kernel-smoothed representation
        density. \\

        $\mathcal{H}[p]$
        & Differential entropy. \\

        $\mathcal{F}[p],E,S$
        & Helmholtz free-energy functional, expected structural potential
        energy, and configuration entropy. \\

        $U(\mathbf{z}_i)$
        & Local structural potential. \\

        $T,T_{\mathrm{eff}}$
        & Macroscopic system temperature and intrinsic hyperspherical
        temperature. \\

        $\mathcal{L}_{\mathrm{align}},
        \mathcal{L}_{\mathrm{unif}},
        \mathcal{L}$
        & Structural alignment loss, uniformity loss, and
        the final training objective. \\
        
        $\kappa_{\mathrm{align}},\kappa$
        &  structural alignment concentration and vMF kernel concentration for global uniformity estimation. \\
        
        $s_{\mathrm{align}},s_{\mathrm{unif}}$
        & Unconstrained log-concentration variables defined as
        $s_{\mathrm{align}}=\log\kappa_{\mathrm{align}}$ and
        $s_{\mathrm{unif}}=\log\kappa$. \\
        \bottomrule
    \end{tabularx}
\end{table}

\section{Theoretical Derivations and Proofs}
\label{app:proofs}

\subsection{Detailed Derivation of the Variational Bridge}
\label{subsection:variational_derivation}

We derive the Kullback--Leibler (KL) identity used in Section~\ref{sec:thermodynamic_variational_bridge}. Fix $T>0$ and a bounded measurable potential $U$ on $\mathbb{S}^{d-1}$. All densities are defined relative to the surface measure $d\sigma$, and $p$ is a probability density with finite differential entropy $\mathcal{H}[p]$. The discrete empirical measure introduced below is not such a density; its entropy is evaluated only after kernel smoothing.

By definition, the KL divergence on the hypersphere $\mathbb{S}^{d-1}$ is given by:
\begin{equation}
    D_{\mathrm{KL}}\bigl(p \parallel p^*\bigr) = \int_{\mathbb{S}^{d-1}} p(\mathbf{z}) \log \frac{p(\mathbf{z})}{p^*(\mathbf{z})} \,d\sigma(\mathbf{z}).
    \label{eq:kl_def}
\end{equation}

For fixed $U$ and $T$, the Gibbs equilibrium density is
\begin{equation}
    p^*(\mathbf{z}) = \frac{1}{\mathcal{Z}} \exp\left( - \frac{U(\mathbf{z})}{T} \right),
\end{equation}
where $\mathcal{Z} = \int_{\mathbb{S}^{d-1}} \exp\bigl( - U(\mathbf{z}) / T \bigr) \,d\sigma(\mathbf{z})$ is the partition function. 

Using the Gibbs density and $\int p\,d\sigma=1$, Equation~\ref{eq:kl_def} becomes
\begin{equation}
    \begin{aligned}
        D_{\mathrm{KL}}(p\parallel p^*)
        &=\int_{\mathbb{S}^{d-1}}p(\mathbf{z})\log p(\mathbf{z})\,d\sigma(\mathbf{z})\\
        &\quad+\frac{1}{T}\int_{\mathbb{S}^{d-1}}p(\mathbf{z})U(\mathbf{z})\,d\sigma(\mathbf{z})+\log\mathcal{Z}\\
        &=-\mathcal{H}[p]+\frac{1}{T}\mathbb{E}_p[U]+\log\mathcal{Z}.
    \end{aligned}
    \label{eq:kl_distributed}
\end{equation}
Defining $\mathcal{F}[p]=\mathbb{E}_p[U]-T\mathcal{H}[p]$ gives
\begin{equation}
    T D_{\mathrm{KL}}(p\parallel p^*)
    =\mathcal{F}[p]+T\log\mathcal{Z}
    =\mathcal{F}[p]-\mathcal{F}[p^*],
\end{equation}
where $\mathcal{F}[p^*]=-T\log\mathcal{Z}$. Nonnegativity of KL divergence implies $\mathcal{F}[p]\geq\mathcal{F}[p^*]$, with equality if and only if $p=p^*$ almost everywhere. This is a fixed-potential, fixed-temperature identity. Since the neighborhood anchors and scale parameters in {\framework} evolve during training, it does not by itself establish convergence of the coupled optimization.

\subsection{Theoretical Analysis: Asymptotic Free-Energy Interpretation}
\label{subsection:theoretical_analysis}

We connect the structural energy and smoothed representation entropy to the encoder objective, yielding an asymptotic free-energy interpretation at fixed scales.

\subsubsection{Notation and Mathematical Preliminaries}

Fix $d\geq2$ and $N$ exactly unit-normalized embeddings $\{\mathbf{z}_i\}_{i=1}^N\subset\mathbb{S}^{d-1}$. Their empirical measure is $\hat{p}=N^{-1}\sum_i\delta_{\mathbf{z}_i}$, where $\delta_{\mathbf{z}_i}$ is the Dirac probability measure. To define differential entropy, we use the von Mises--Fisher (vMF) kernel density estimate
\begin{equation}
    \hat{p}_\kappa(\mathbf{z})=\frac{C_d(\kappa)}{N}\sum_{i=1}^N\exp(\kappa\mathbf{z}_i^\top\mathbf{z}),
\end{equation}
with normalization constant
\begin{equation}
    C_d(\kappa)=\frac{\kappa^{d/2-1}}{(2\pi)^{d/2}I_{d/2-1}(\kappa)},
\end{equation}
where $I_v$ is the modified Bessel function of the first kind. The concentration $\kappa>0$ is held fixed when varying embeddings; the asymptotic analysis considers $\kappa\to0^+$ at fixed $d$.

\subsubsection{Structural Potential Energy Mapping}
\label{subsubsection:potential_energy_mapping}

For a configuration $\mathbf{Z}$ with unit neighborhood anchors, define the node-dependent potential
\begin{equation}
    U_i(\mathbf{z};\mathbf{Z})
    =w_i\bigl(1-\langle\mathbf{z},\boldsymbol{\mu}_i(\mathbf{Z})\rangle\bigr),
\end{equation}
where $w_i=[\sigma(|\mathcal{N}_i|)]^\tau$, $\sigma$ is the sigmoid function, and $\boldsymbol{\mu}_i$ is defined in Section~\ref{sec:structural_binding_energy}. We assume nonempty neighborhoods with nonzero mean representations; $U_i$ depends on both the node index and $\mathbf{Z}$.

\begin{proposition}[Empirical Structural Energy]
\label{prop:potential_energy_equivalence}
The mean node-wise structural potential equals the alignment loss:
\begin{equation}
\begin{aligned}
    E(\mathbf{Z})
    &\triangleq\frac{1}{N}\sum_{i=1}^N U_i(\mathbf{z}_i;\mathbf{Z})\\
    &=\frac{1}{N}\sum_{i=1}^N w_i\bigl(1-\langle\mathbf{z}_i,\boldsymbol{\mu}_i\rangle\bigr)
    =\mathcal{L}_{\mathrm{align}}.
\end{aligned}
\end{equation}
\end{proposition}

\begin{proof}
Substituting $U_i(\mathbf{z}_i;\mathbf{Z})=w_i(1-\langle\mathbf{z}_i,\boldsymbol{\mu}_i\rangle)$ into the node average recovers the definition of $\mathcal{L}_{\mathrm{align}}$.
\end{proof}

For unit embeddings and anchors, $1-\langle\mathbf{z}_i,\boldsymbol{\mu}_i\rangle=\tfrac12\|\mathbf{z}_i-\boldsymbol{\mu}_i\|_2^2$, giving a squared-distance interpretation with configuration-dependent anchors.

\subsubsection{Asymptotic Hyperspherical Entropy Expansion}
\label{subsubsection:entropy_uniformity_duality}

We next relate the first-moment dispersion loss to the leading representation-dependent term in the entropy of the vMF-smoothed density.

\begin{proposition}[Asymptotic Entropy--Dispersion Relation]
\label{prop:entropy_uniformity}
For fixed $d\geq2$ and $N$ unit embeddings, the vMF KDE defined above satisfies, as $\kappa\to0^+$,
\begin{equation}
    \mathcal{H}[\hat{p}_\kappa] = \log|\mathbb{S}^{d-1}| - \frac{\kappa^2}{2d}\mathcal{L}_{\mathrm{unif}} + O(\kappa^3).
\end{equation}
Consequently, the first-moment dispersion loss is related to the entropy deficit by
\begin{equation}
    \mathcal{L}_{\mathrm{unif}} = \frac{2d}{\kappa^2}\bigl(\log|\mathbb{S}^{d-1}| - \mathcal{H}[\hat{p}_\kappa]\bigr)+O(\kappa).
\end{equation}
\end{proposition}

\begin{proof}
Let $d\bar{\sigma}=d\sigma/|\mathbb{S}^{d-1}|$ be the uniform probability measure and $\tilde{p}_\kappa=|\mathbb{S}^{d-1}|\hat{p}_\kappa$ the density relative to it. The differential entropy is
\begin{equation}
    \mathcal{H}[\hat{p}_\kappa] = \log|\mathbb{S}^{d-1}| - \int_{\mathbb{S}^{d-1}} \tilde{p}_\kappa(\mathbf{z})\log\tilde{p}_\kappa(\mathbf{z})\,d\bar{\sigma}(\mathbf{z}).
\end{equation}
Using $|\mathbb{S}^{d-1}|C_d(\kappa)=1-\frac{\kappa^2}{2d}+O(\kappa^4)$, the second-order density expansion is
\begin{equation}
    \begin{aligned}
        \tilde{p}_\kappa(\mathbf{z}) &= \left(1 - \frac{\kappa^2}{2d}\right)\frac{1}{N}\sum_{i=1}^N \left(1 + \kappa\mathbf{z}_i^\top\mathbf{z} + \frac{\kappa^2}{2}(\mathbf{z}_i^\top\mathbf{z})^2\right) + O(\kappa^3) \\
        &= 1 + \kappa\bar{\mathbf{z}}^\top\mathbf{z} + \frac{\kappa^2}{2}\left[\frac{1}{N}\sum_{i=1}^N(\mathbf{z}_i^\top\mathbf{z})^2 - \frac{1}{d}\right] + O(\kappa^3),
    \end{aligned}
\end{equation}
where $\bar{\mathbf{z}} = \frac{1}{N}\sum_{i=1}^N \mathbf{z}_i$ is the empirical mean vector. Since $|\mathbf{z}_i^\top\mathbf{z}|\leq1$, the Taylor remainders are uniform over the sphere and unit-embedding configurations at fixed $d$, which justifies term-wise integration.

Using the Taylor expansion $(1+x)\log(1+x) = x + \frac{1}{2}x^2 + O(x^3)$, we substitute $x = \tilde{p}_\kappa(\mathbf{z}) - 1$ into the integral:
\begin{equation}
    \begin{aligned}
        \tilde{p}_\kappa(\mathbf{z})\log\tilde{p}_\kappa(\mathbf{z}) ={}& \kappa\bar{\mathbf{z}}^\top\mathbf{z} + \frac{\kappa^2}{2}\left[\frac{1}{N}\sum_{i=1}^N(\mathbf{z}_i^\top\mathbf{z})^2 - \frac{1}{d}\right] \\
        &+ \frac{1}{2}\left(\kappa\bar{\mathbf{z}}^\top\mathbf{z}\right)^2 + O(\kappa^3).
    \end{aligned}
\end{equation}
The spherical moment identities
\begin{equation}
    \int\mathbf{z}\,d\bar{\sigma}=\mathbf{0},
    \qquad \int(\mathbf{v}^\top\mathbf{z})^2\,d\bar{\sigma}=\frac{\|\mathbf{v}\|^2}{d}
\end{equation}
hold for any $\mathbf{v}\in\mathbb{R}^d$. Applying them to $\mathbf{z}_i$ and $\bar{\mathbf{z}}$ cancels the first two terms above.
Consequently, only the squared first-order term contributes at second order:
\begin{equation}
    \begin{aligned}
        \int_{\mathbb{S}^{d-1}} \tilde{p}_\kappa(\mathbf{z})\log\tilde{p}_\kappa(\mathbf{z})\,d\bar{\sigma}(\mathbf{z}) &= \frac{\kappa^2}{2d}\|\bar{\mathbf{z}}\|^2 + O(\kappa^3) \\
        &= \frac{\kappa^2}{2d}\mathcal{L}_{\mathrm{unif}} + O(\kappa^3).
    \end{aligned}
\end{equation}
Substituting into the entropy identity proves the first expansion. Multiplication by $2d/\kappa^2$ changes the $O(\kappa^3)$ remainder to $O(\kappa)$ at fixed $d$, yielding the second relation.
\end{proof}

The expansion requires small $\kappa$ at fixed $d$; high dimensionality alone is insufficient. Moreover, $\mathcal{L}_{\mathrm{unif}}$ controls only the first moment: equally weighted antipodal embeddings have zero mean without being uniform, so higher-order entropy terms still matter.

\subsubsection{Asymptotic Free-Energy Correspondence}
\label{subsubsection:free_energy_minimization}

Combining the energy identity and entropy expansion gives the following correspondence for fixed coefficients.

\begin{theorem}[Asymptotic Free-Energy Correspondence]
\label{theorem:helmholtz_free_energy_equivalence}
Fix $d\geq2$, $N$, and $\alpha>0$. For each sufficiently small $\kappa>0$, define the configuration-dependent free-energy surrogate
\begin{equation}
    \mathcal{F}_\kappa(\mathbf{Z})
    =E(\mathbf{Z})-T_\kappa\mathcal{H}[\hat{p}_\kappa],
\end{equation}
with $E(\mathbf{Z})=\mathcal{L}_{\mathrm{align}}$ and
\begin{equation}
    T_\kappa=\frac{\alpha}{T_{\mathrm{eff}}}
    =\alpha\frac{2d}{\kappa^2},
    \qquad T_{\mathrm{eff}}=\frac{\kappa^2}{2d}.
\end{equation}
As $\kappa\to0^+$, the fixed-coefficient encoder objective satisfies
\begin{equation}
\begin{aligned}
    \mathcal{L}_\alpha
    &\triangleq\mathcal{L}_{\mathrm{align}}+\alpha\mathcal{L}_{\mathrm{unif}}\\
    &=\mathcal{F}_\kappa(\mathbf{Z})
    +T_\kappa\log|\mathbb{S}^{d-1}|+O(\kappa).
\end{aligned}
\end{equation}
The additive term is independent of $\mathbf{Z}$ for fixed $\alpha$ and $\kappa$. The result is an asymptotic objective correspondence, not an exact equality of minimizers or a convergence guarantee.
\end{theorem}

\begin{proof}
Substituting Propositions~\ref{prop:potential_energy_equivalence} and~\ref{prop:entropy_uniformity} into the encoder objective yields
\begin{equation}
\begin{aligned}
    \mathcal{L}_\alpha
    &=E(\mathbf{Z})+\alpha\mathcal{L}_{\mathrm{unif}}\\
    &=E(\mathbf{Z})+\frac{2d\alpha}{\kappa^2}
    \bigl(\log|\mathbb{S}^{d-1}|-\mathcal{H}[\hat{p}_\kappa]\bigr)\\
    &\quad+O(\alpha\kappa)\\
    &=E(\mathbf{Z})-T_\kappa\mathcal{H}[\hat{p}_\kappa]
    +T_\kappa\log|\mathbb{S}^{d-1}|+O(\kappa),
\end{aligned}
\end{equation}
where the last remainder uses fixed $\alpha$. The definition of $\mathcal{F}_\kappa$ completes the proof.
\end{proof}

Here, $\mathcal{F}_\kappa$ combines empirical node-wise energy with KDE entropy, so it differs from the same-density Gibbs functional in Appendix~\ref{subsection:variational_derivation}.

With scales fixed, Equation~\ref{eq:bayesian_thermostat} has the form $a\mathcal{L}_{\mathrm{align}}+b\mathcal{L}_{\mathrm{unif}}+r$, with $a=\kappa_{\mathrm{align}}$, $b=d/\kappa^2$, and $r$ independent of the encoder parameters. Dividing by $a>0$ gives $\alpha=b/a$. During joint optimization, these quantities and $T_\kappa\log|\mathbb{S}^{d-1}|$ vary, and the remainder retains its $\alpha$ dependence. Thus, the fixed-coefficient result does not establish free-energy equivalence or convergence for the full encoder--thermostat dynamics.

\section{Experimental Details}
\label{app:experiments}

\subsection{Datasets}
\label{app:datasets}

Table~\ref{tab:dataset_stat} summarizes the statistics of the eleven benchmarks.

\begin{table}[H]
\centering
\caption{Statistics of the benchmark datasets.}
\label{tab:dataset_stat}
\small
\setlength{\tabcolsep}{4pt}
\begin{tabular}{l|c|rrrr}
\toprule
Dataset & Type & Nodes & Edges & Features & Classes \\
\midrule
Cora & \multirow{7}{*}{Homophilic} & 2,708 & 5,278 & 1,433 & 7 \\
CiteSeer & & 3,327 & 4,522 & 3,703 & 6 \\
PubMed & & 19,717 & 44,324 & 500 & 3 \\
Computers & & 13,752 & 245,861 & 767 & 10 \\
Photo & & 7,650 & 119,081 & 745 & 8 \\
CS & & 18,333 & 81,894 & 6,805 & 15 \\
Physics & & 34,493 & 247,962 & 8,415 & 5 \\
\midrule
Actor & \multirow{4}{*}{Heterophilic} & 7,600 & 30,019 & 932 & 5 \\
Cornell & & 183 & 298 & 1,703 & 5 \\
Texas & & 183 & 325 & 1,703 & 5 \\
Wisconsin & & 251 & 515 & 1,703 & 5 \\
\bottomrule
\end{tabular}
\end{table}

\subsection{Evaluation Protocols}
\label{app:evaluation}

For node classification, we follow the linear evaluation protocol of \citep{sun2024rethinking}. 
Specifically, the graph encoder is pretrained in a self-supervised manner and then frozen, and an $\ell_2$-regularized logistic regression classifier is trained on the learned embeddings.
Nodes are randomly partitioned into training/validation/testing sets with ratios of 10\%/10\%/80\% for homophilic graphs and 48\%/32\%/20\% for heterophilic graphs.

For link prediction, we follow~\citep{li2023evaluating}, where the embeddings of node pairs are concatenated and fed into a two-layer MLP decoder to predict missing edges. 
The observed edges are randomly partitioned into 85\%/5\%/10\% for training, validation, and testing, respectively. 

\subsection{Implementation Details}
\label{app:implementation}

We implement {\framework} using the PyTorch Geometric library. The encoder comprises a single TransformerConv layer followed by a SiLU activation. Unless otherwise specified, the embedding dimension is set to $d=1024$, with $\tau=5$, $p_e=0.8$, and $p_x=0.1$ across all datasets.
The Bayesian thermostat is initialized with $\kappa_{\mathrm{task}}=d$, while $\kappa$ is initialized to $1.0$ for homophilic graphs and $0.2$ for heterophilic graphs.

The encoder and the trainable thermostat parameters are jointly optimized using Adam with a learning rate of $10^{-3}$. A weight decay of $10^{-5}$ is applied to the encoder parameters, whereas no weight decay is applied to the thermostat parameters.
Training is conducted for up to 1500 epochs with early stopping based on the self-supervised training loss.
All experiments are conducted on a single NVIDIA RTX 3090 GPU. 
For downstream evaluation, the pretrained encoder remains frozen to ensure a fair and consistent assessment of the learned representations.

\section{Additional Experiments}
\label{app:additional_experiments}

\subsection{Representation Visualization}
\label{app:representation_visualization}

\begin{figure}[!htbp]
    \centering
    \subfloat[Raw Feature]{%
        \includegraphics[trim={2mm 2mm 2mm 1mm}, clip, width=0.30\columnwidth]{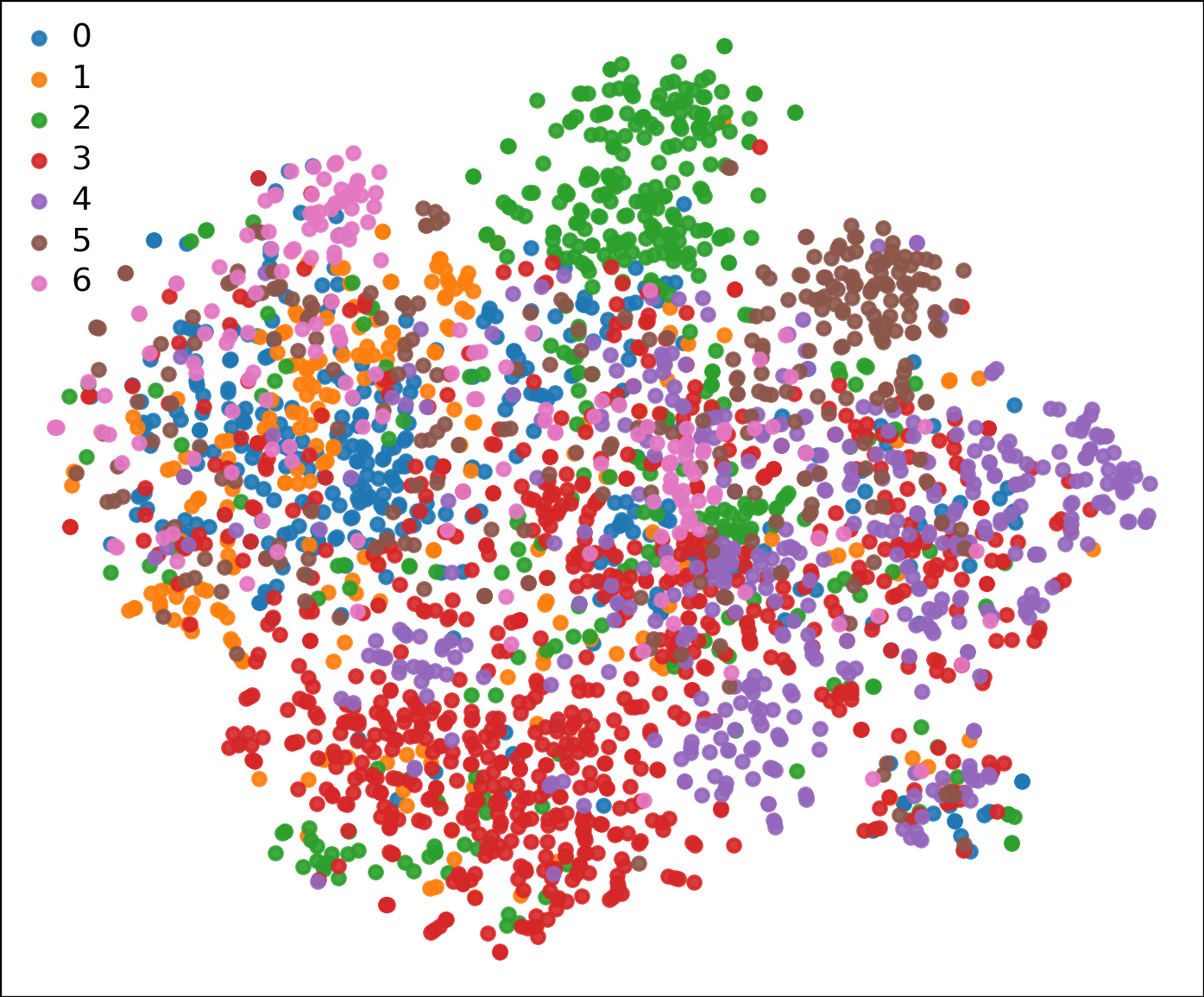}
    }\hspace{0.01\columnwidth}
        \subfloat[GRACE]{%
        \includegraphics[trim={2mm 2mm 2mm 1mm}, clip, width=0.30\columnwidth]{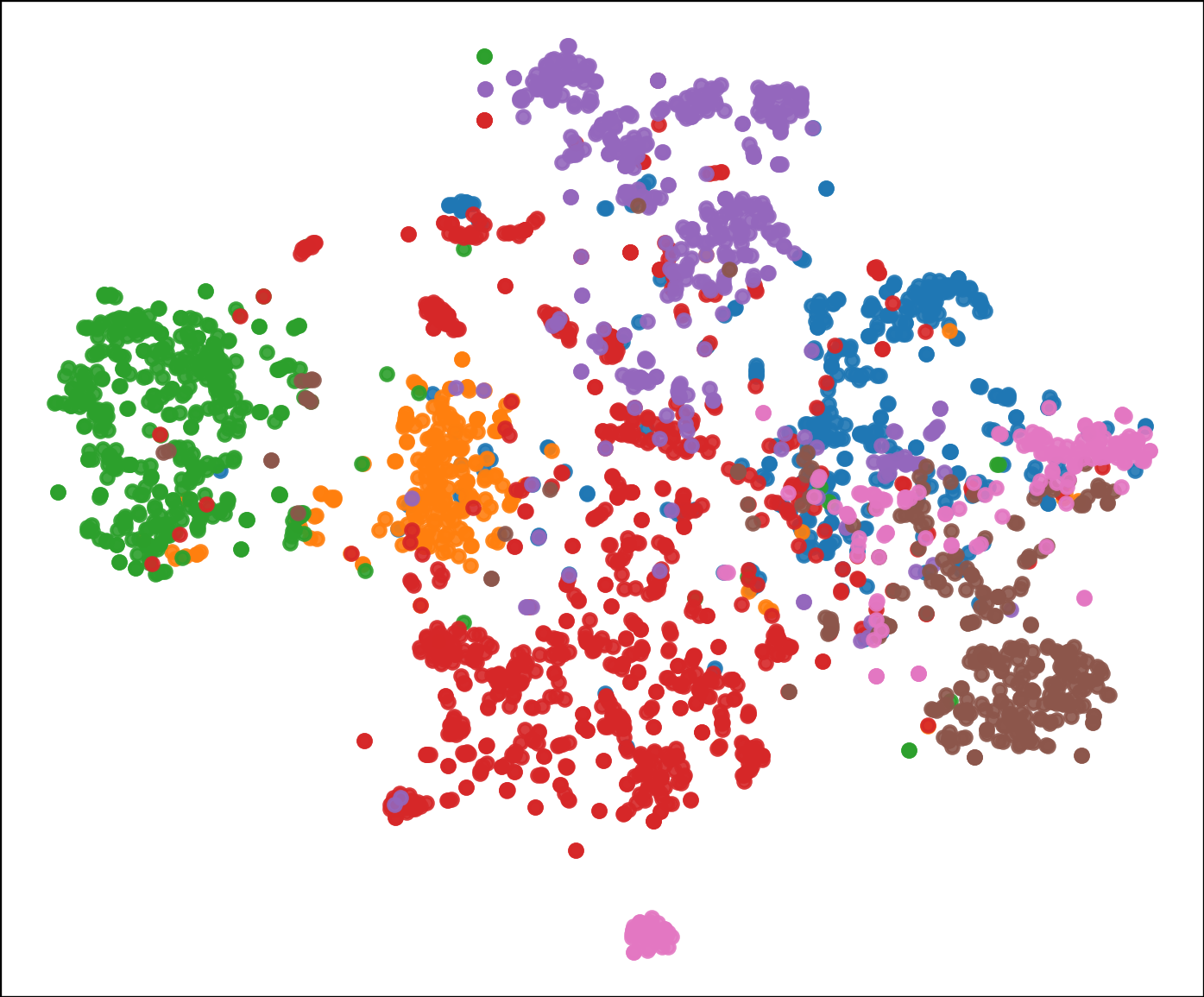}
    }\hspace{0.01\columnwidth}
        \subfloat[DGI]{%
        \includegraphics[trim={2mm 2mm 2mm 1mm}, clip, width=0.30\columnwidth]{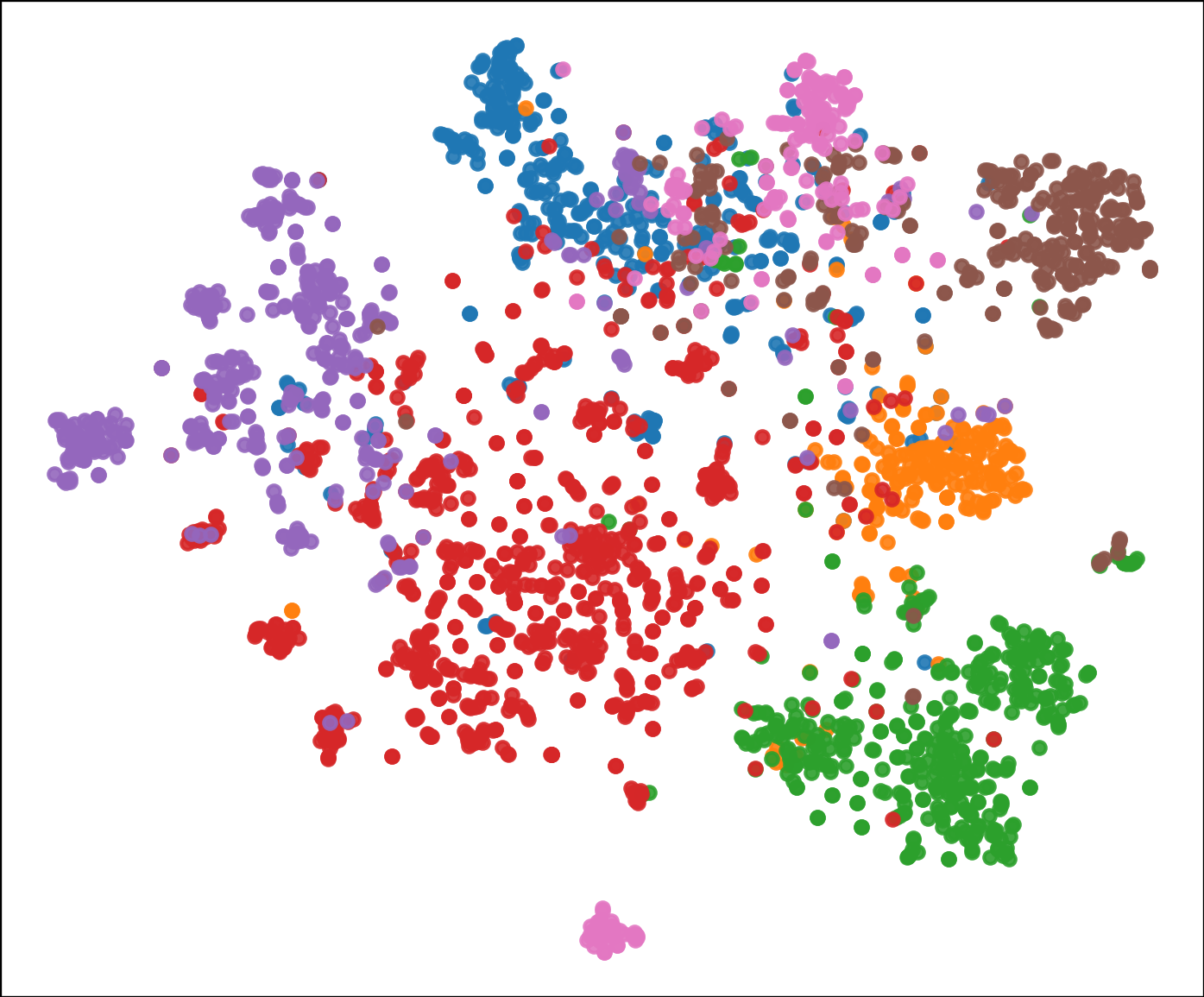}
    }\hspace{0.01\columnwidth}
    \subfloat[BGRL]{%
        \includegraphics[trim={2mm 2mm 2mm 1mm}, clip, width=0.30\columnwidth]{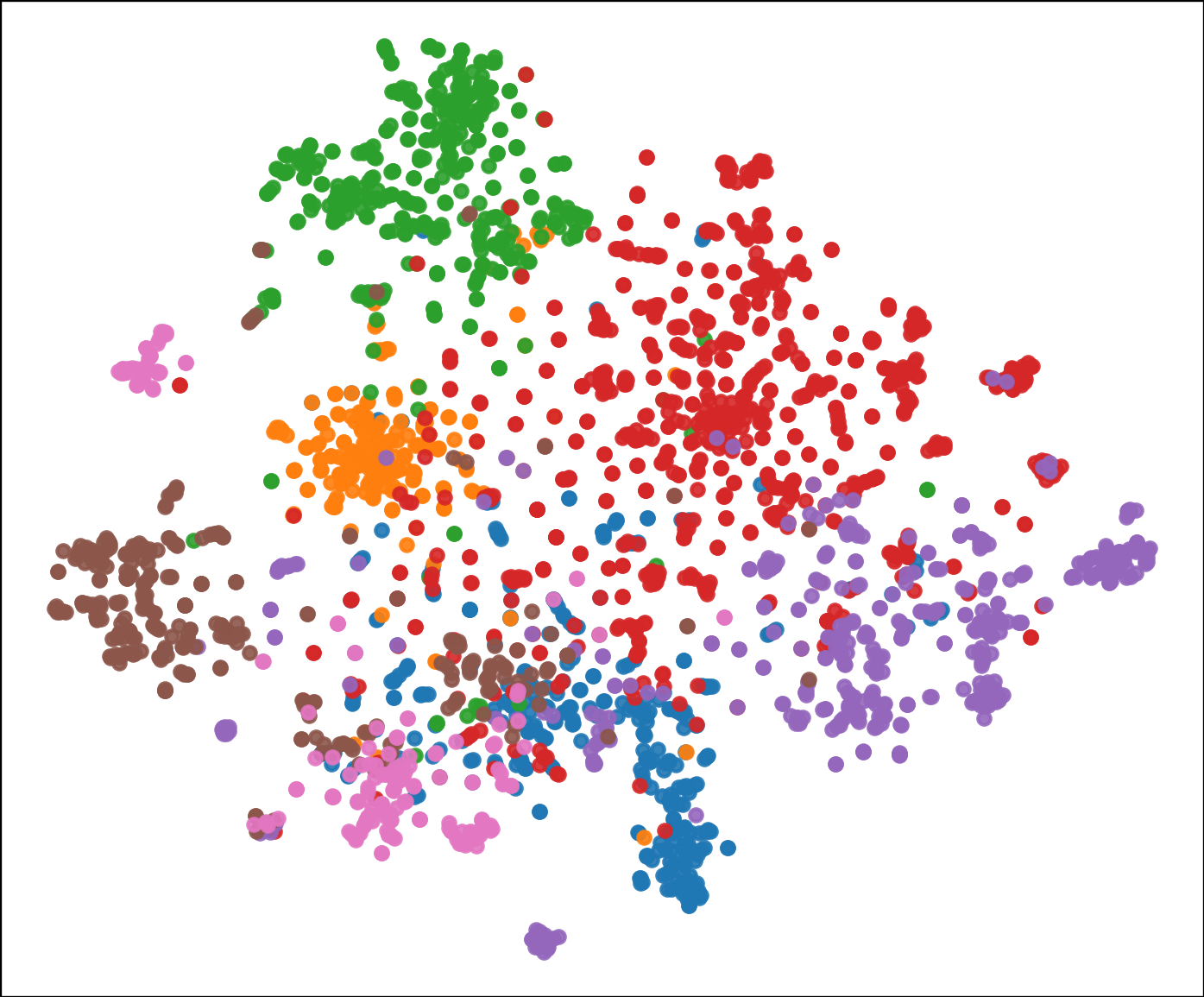}
    }\hspace{0.01\columnwidth}
    \subfloat[SGRL]{%
        \includegraphics[trim={2mm 2mm 2mm 1mm}, clip, width=0.30\columnwidth]{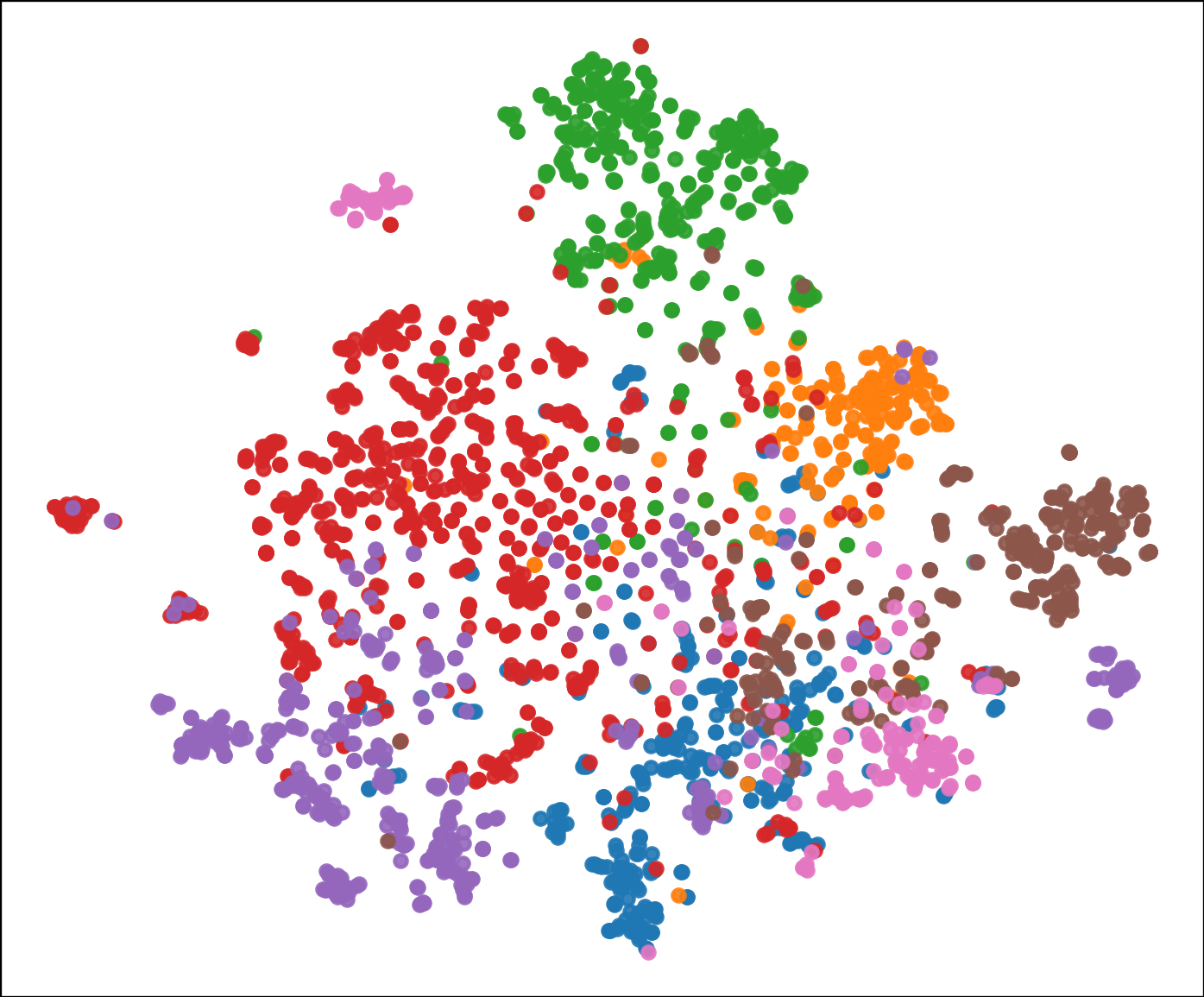}
    }\hspace{0.01\columnwidth}
    \subfloat[\framework]{%
        \includegraphics[trim={21mm 16mm 21mm 17mm}, clip, width=0.30\columnwidth]{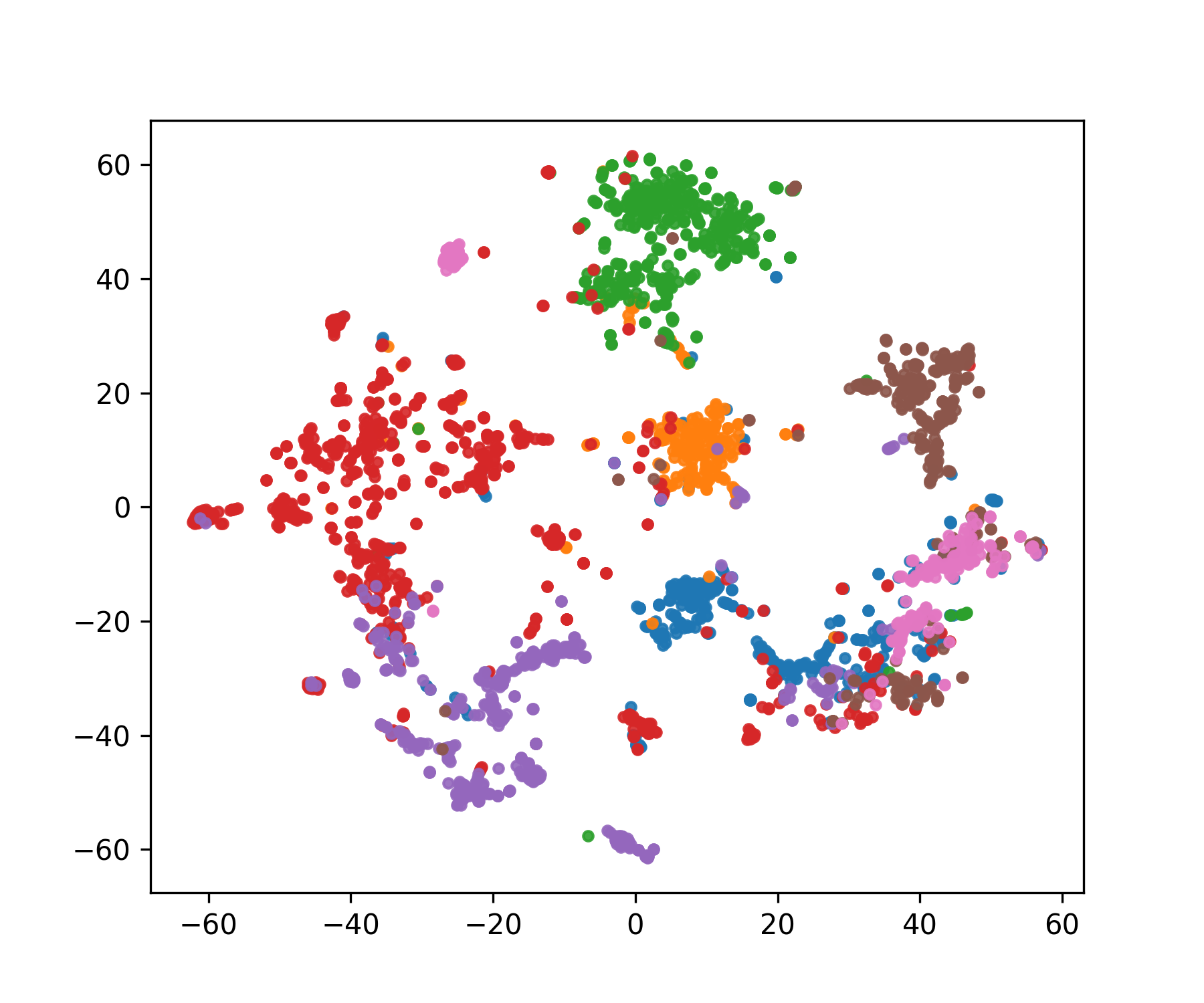}
    }
    \caption{t-SNE embeddings of nodes in the Cora dataset.}
    \label{fig:tsne_cora}
\end{figure}

Fig.~\ref{fig:tsne_cora} provides the embedding visualizations discussed in Section~\ref{sec:rq3}.

\subsection{Robustness and Over-smoothing Resistance}
\label{app:robustness}

\begin{figure}[!htbp]
    \centering
    \subfloat[Hidden dimension (Cora)\label{fig:dim_cora}]{%
        \includegraphics[width=0.49\columnwidth]{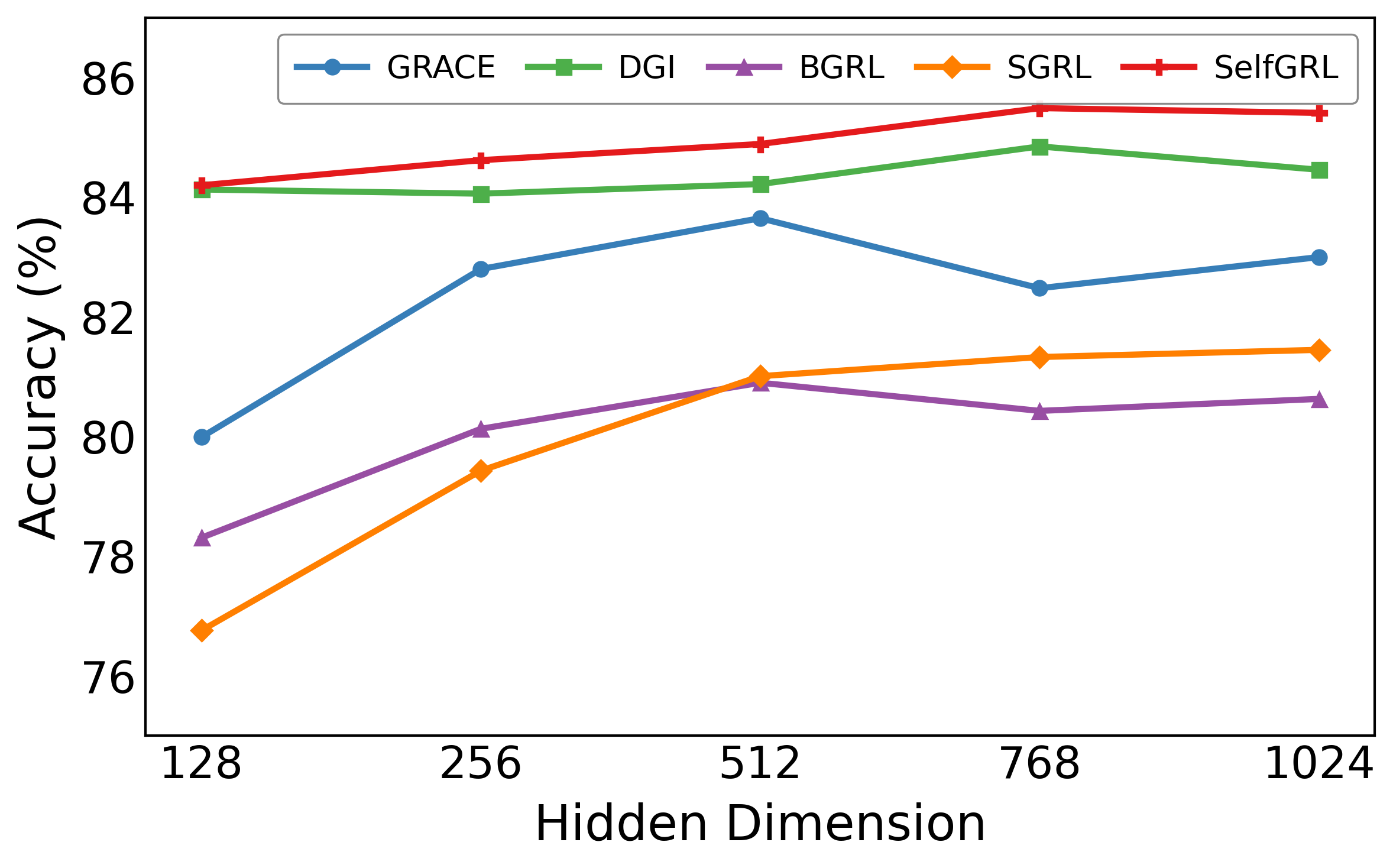}
    }\hfill
    \subfloat[Hidden dimension (Texas)\label{fig:dim_texas}]{%
        \includegraphics[width=0.49\columnwidth]{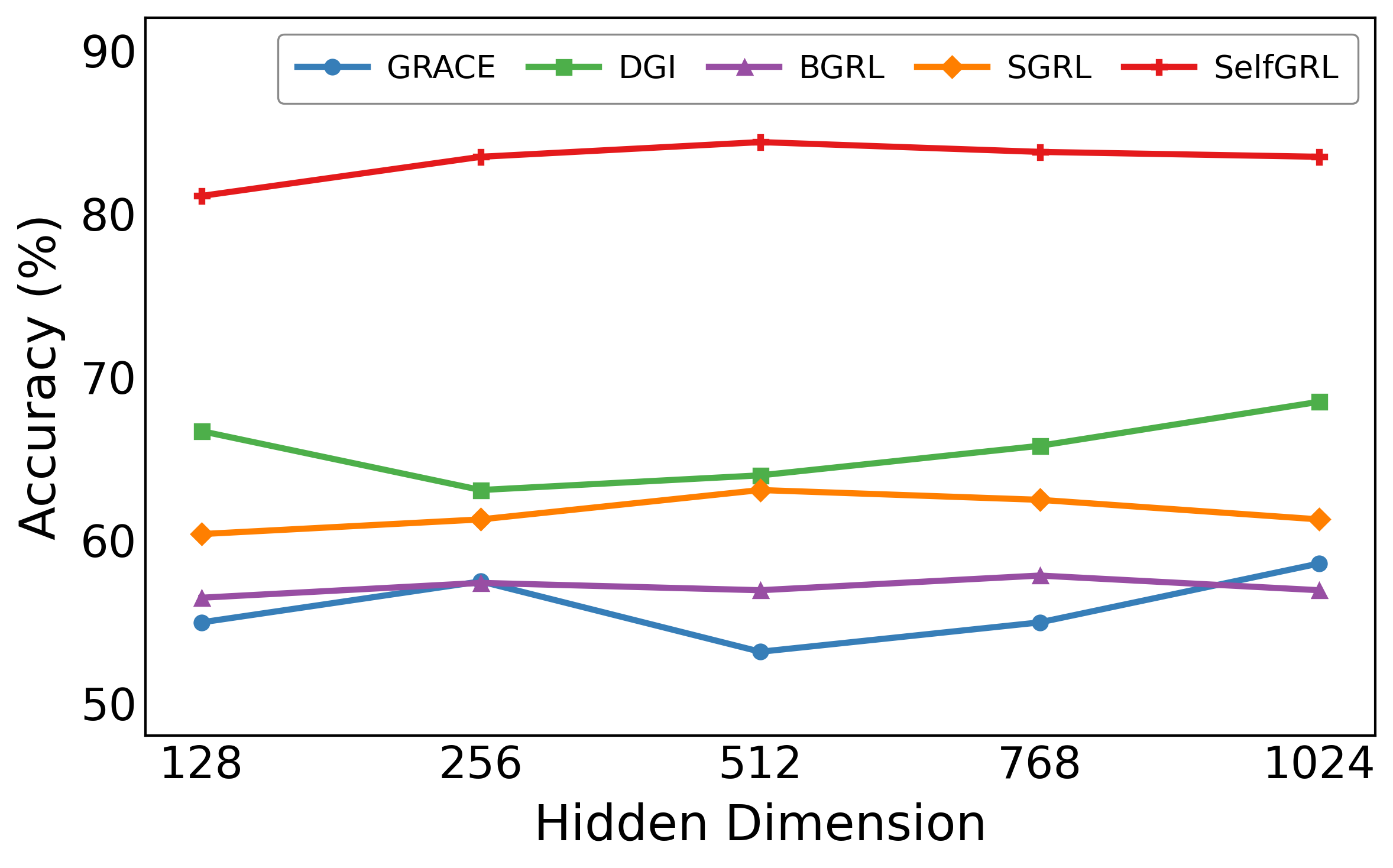}
    }\\
    \subfloat[Network depth (Cora)\label{fig:depth_cora}]{%
        \includegraphics[width=0.49\columnwidth]{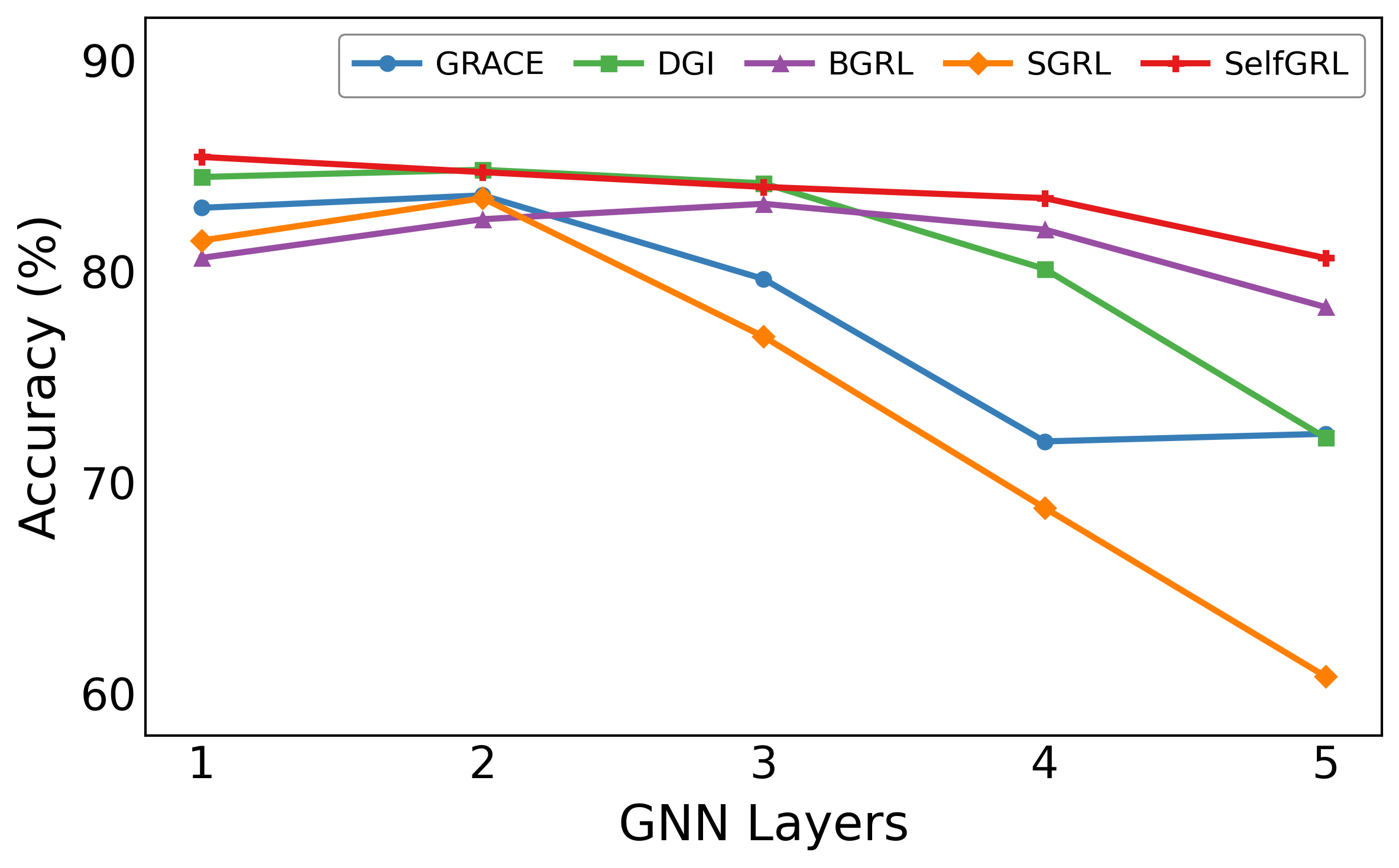}
    }\hfill
    \subfloat[Network depth (Texas)\label{fig:depth_texas}]{%
        \includegraphics[width=0.49\columnwidth]{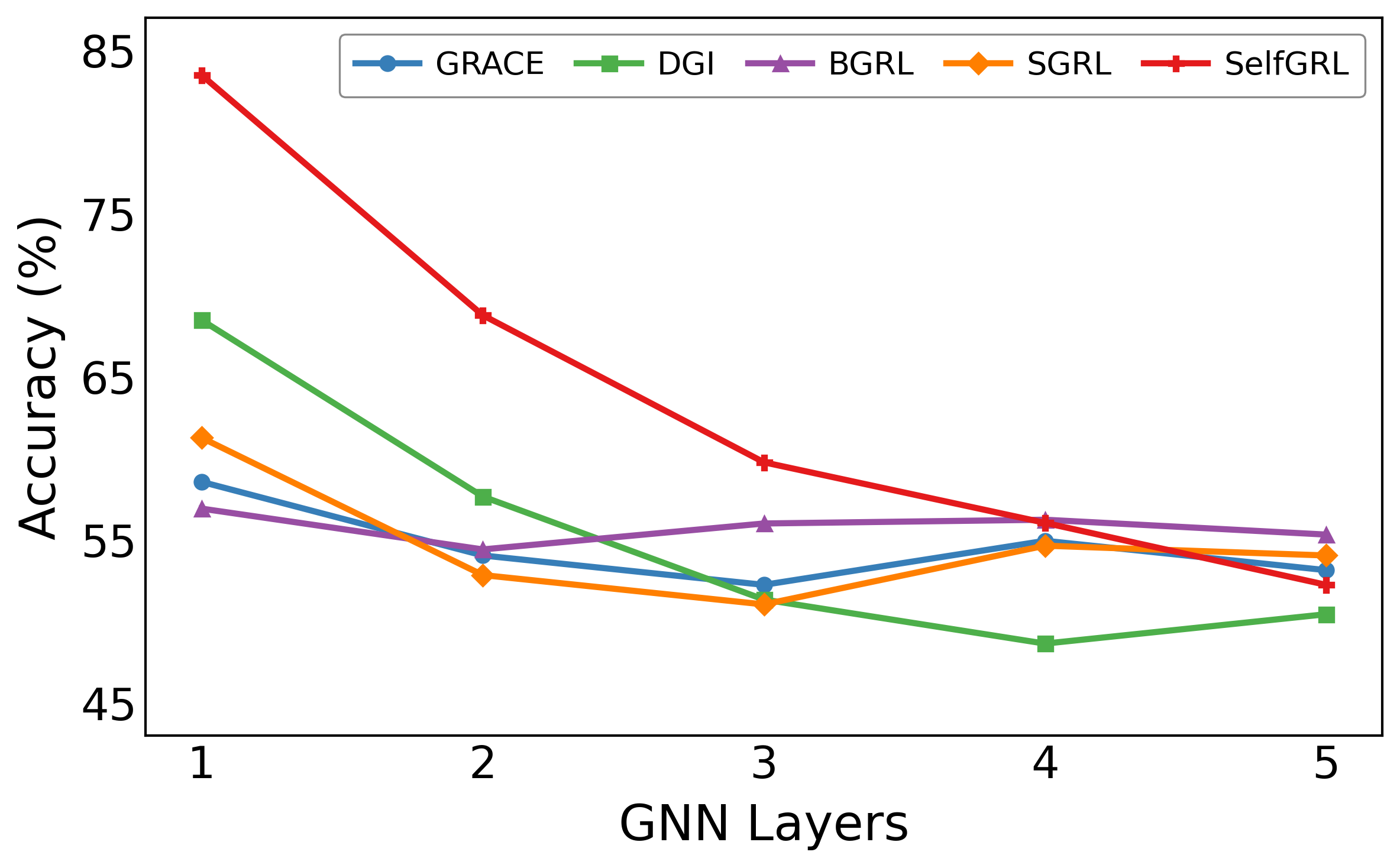}
    }
    \caption{Node classification accuracy (\%) with varying hidden dimensions (a--b) and network depths (c--d).}
    \label{fig:robustness}
\end{figure}

Fig.~\ref{fig:robustness}(a--b) evaluates robustness to hidden dimension on Cora and Texas using a single-layer encoder, with dimensions ranging from $128$ to $1024$. {\framework} achieves the highest accuracy across all evaluated dimensions on both datasets, with modest performance variation and a substantial advantage on Texas. Its accuracy rises slightly before stabilizing, whereas some baselines exhibit more noticeable fluctuations, particularly GRACE and SGRL on Texas. These results suggest that {\framework} learns effective representations across a broad range of embedding dimensions without dimension-specific tuning.

Fig.~\ref{fig:robustness}(c--d) examines the effect of increasing encoder depth from one to five layers with the hidden dimension fixed at $1024$. On Cora, {\framework} performs best at four and five layers, retaining accuracy above $80\%$ at five layers while several baselines decline more sharply. On Texas, it leads at one to three layers, but its performance deteriorates with further depth and its advantage diminishes. These results suggest improved tolerance to deeper propagation on Cora, while highlighting the remaining challenges of deep representation learning under strong heterophily.

\end{document}